\documentclass[11pt,a4paper]{article}
\usepackage[left=3cm, right=3cm, top=3.5cm]{geometry}
\usepackage{amssymb,amsmath, amsthm, amsfonts, rotating}
\usepackage{mathrsfs,color,bbold,url}
\usepackage{stmaryrd}
\usepackage{tikz}
\usepackage[all]{xypic}
\usepackage{enumerate}

\newtheorem{theorem}{Theorem}[section]
\newtheorem{lemma}[theorem]{Lemma}
\newtheorem{corollary}[theorem]{Corollary}
\newtheorem{proposition}[theorem]{Proposition}

\theoremstyle{definition}
\newtheorem{definition}[theorem]{Definition}

\newtheorem{example}[theorem]{Example}

\usepackage{marvosym}
\usepackage{nicefrac}
\usepackage{circuitikz}
\usepackage{adjustbox}
\usepackage[most]{tcolorbox}

\usepackage{tikz-qtree}
\usepackage{tikz-qtree-compat}

\tikzset{every tree node/.style={align=center,anchor=north}}

\makeatother

\begin{document}

\title{The Horn Non-Clausal Class and its Polynomiality}
\author{Gonzalo E. Imaz\\ [2mm] %
{\small  Artificial Intelligence Research Institute (IIIA) - CSIC, Barcelona, Spain}\\
{\small \tt email: {\{gonzalo\}}@iiia.csic.es}\\[1mm]
}

\date{}
\maketitle

\begin{abstract}
    
The expressiveness      of     propositional non-clausal (NC)
 formulas   is exponentially  richer  than that
    of   clausal formulas.  Yet,  clausal efficiency    outperforms
       non-clausal efficiency. Indeed,   a major weakness of the latter  is that, while    
       Horn clausal formulas,  along with     Horn  algorithms,  
       are  crucial for the  high  efficiency of clausal  reasoning,  
       {no  Horn-like formulas in non-clausal form had been proposed.}
 
To overcome such weakness,  we  first define  {\em  the hybrid class $\mathbb{H_{NC}}$} of  
 {\em  Horn Non-Clausal  (Horn-NC) formulas}    by   adequately 
  lifting the  Horn   pattern  to     NC form, and
  argue that  $\mathbb{H_{NC}}$, along with  future   Horn-NC algorithms  
 shall  increase    non-clausal   efficiency  
     just as the Horn class has increased      clausal efficiency.
   
    Secondly,  we:  (i)  give  the  
      compact,  inductive definition of $\mathbb{H_{NC}}$;    
    (ii)  prove that syntactically $\mathbb{H_{NC}}$ subsumes the Horn class but
    semantically both classes are  equivalent;
    and (iii) characterize 
     the non-clausal formulas  belonging to $\mathbb{H_{NC}}$.    
     
Thirdly, we   define the {\em   Non-Clausal  Unit-Resolution} calculus, or  UR$_{\mbox{nc}}$, and prove that
it   checks the satisfiability of $\mathbb{H_{NC}}$ in polynomial time.
This fact,
 to our knowledge,  makes {\em $\mathbb{H_{NC}}$ the first
characterized  polynomial  class in NC reasoning.}


Finally, we prove   that  $\mathbb{H_{NC}}$ is  linearly recognizable, and also that 
 it is both strictly more succinct  and exponentially richer than the Horn class.

 We discuss that  in NC automated reasoning, e.g. satisfiability solving,    theorem
proving,   logic programming, etc.,  can directly benefit from $\mathbb{H_{NC}}$ and 
UR$_{\mbox{nc}}$ and that, 
as a by-product of its proved   properties,
  $\mathbb{H_{NC}}$  arises as a new alternative  to analyze Horn  functions and implication systems.

\vspace{.15cm}   
\noindent  {\bf Field:}   {\em Automated Reasoning  in Propositional    Logic.}
 
\vspace{.05cm} 
\end{abstract}

\newenvironment{niceproof}{\trivlist\item[\hskip
\labelsep{\it Proof.\/}]\ignorespaces}{\hfill$\blacksquare$\endtrivlist}

\newenvironment{proofsketch}{\trivlist\item[\hskip
\labelsep{\it Proof Sketch.\/}]\ignorespaces}{\hfill$\blacksquare$\endtrivlist}

\newenvironment{niceproofsketch}{\trivlist\item[\hskip
       \labelsep{\bf Proof Sketch.\/}]\ignorespaces}{\hfill$\blacksquare$\endtrivlist}



\section{Introduction}


\vspace{.05cm}
The   expressive power of   {\em  non-clausal (NC)    formulas}   
is exponentially  richer (formally specified in Section \ref{sec:semanticalsyntactical}) than that of clausal formulas  
and  they have 
 found much use in heterogeneous fields and  practical settings     such as
 diagnosis  \cite{DBLP:journals/jair/Darwiche98},
 algebric theorems \cite{MeierSorge05}, linear monadic decomposition  \cite{HagueLRW20},
  symbolic model checking \cite{WilliamsBiereClarkeGupta00},
  formal verification \cite{DsilvaKroeningWeissenbacher08},
supervisory control \cite{ClaessenEenSheeranSorenssonVoronvAkesson09},   circuit
verification \cite{DrechslerJunttilaNiemela09},   linear quantifier elimination \cite{Nipkow10},
quantified boolean formulas \cite{EglySW09},
constraint  problems \cite{VekslerS16},   
theorem proving \cite{Otten11},
   ontologies \cite{DBLP:journals/logcom/AlvezHLR19},   
 knowledge compilation
\cite{DBLP:journals/jacm/Darwiche01}, 
    heuristics \cite{DBLP:journals/entcs/BarrettD05},  
satisfiability modulo theory \cite{DBLP:conf/cav/JhaLS09}, 
MaxSAT \cite{LiMS19},  
 encodings \cite{NavarroVoronkov05},  
 modal logics \cite{DBLP:journals/jar/GaintzarainHLNO13}, 
nested expressions   \cite{DBLP:journals/tplp/PearceTW09}, 
description logics \cite{DBLP:journals/jar/KlarmanES11}, many-valued logics \cite{00010HZC18}, 
learning \cite{AbouziedAPHS13}, deductive databases \cite{DohertyKS99}, 
minimum unsatisfiability \cite{BuningKullmann09}, compiling linear constrains   \cite{TamuraTKB09}, 
 constraint handling rules \cite{GiustoGM12},  
dynamic  systems  \cite{Platzer08},  DPLL \cite{ThiffaultBacchusWalsh04},
 linear constraint solving \cite{AudemardBCKS02b}, 
linear arithmetic games \cite{FarzanK18}, aggregates \cite{CabalarFSS19},
equilibrium logic \cite{Pearce06}, model generation   \cite{JacksonS08}, 
constrained Horn clauses \cite{KomuravelliBGM15}, Horn clause verification \cite{UnnoT15},   constraint logic
programs \cite{abs-2007-03656},  NC   resolution \cite{Murray82}, 
Skolemisation \cite{KotelnikovK0V16}, numeric planning \cite{ScalaHTR20},   
stochastic search \cite{MuhammadS06} and   hard problems \cite{NavarroVoronkov05}.

\vspace{.05cm}
On the other side,   the     {\it  Horn clausal  formulas }are  pivotal in our approach. 
Such formulas can be read naturally as 
 instructions for a computer and are central for  
deductive databases, declarative 
programming,        automated reasoning and artificial intelligence.
Modeling and computing   with Horn languages  have received  a great deal of attention   
since 1943 \cite{McKinsey43,Horn51}  and  they are the core of  countless  contemporary   research  fields
spread across many logics and a variety of reasoning problems.

\vspace{.05cm}
Regarding  specifically  the conjunctive r\^ole of      Horn formulas and  Horn-SAT algorithms in propositional logic, 
 its linearity   \cite{DowlingGallier84,ItaiMakowsky87,Minoux88,Scutella90,GhallabEscalada91, DuboisAndreBoufkhadCarlier96}
   for  both satisfiability  checking     Horn  formulas   and   
  simplifying    non-Horn ones is  likely its most   significant, well-known  virtue.   
The valuable contribution  of    Horn   efficiency to the global efficiency of clausal reasoning
is evidenced  by the fact that  highly-efficient  DPLL solvers embed a  
 Horn-SAT-like algorithm
so-called   {\em Unit Propagation}\footnote{So the terms 
    {\em Horn-SAT  algorithm} and   {\em Unit Propagation} will be used interchangeably.}\cite{DixonGP04}. 
%
     Hence, finding   polynomial  clausal 
    super-classes of the Horn class has been  
a key issue for several decades in the quest for improving clausal reasoning; some found classes are:
hidden Horn \cite{Lewis78,Aspvall80}, generalized  Horn    \cite{ChandruCoullardHammerMontanezSun90},
    Q-Horn    \cite{BorosCramaHammer90},     extended-Horn  \cite{ChandruHooker91}, SLUR     \cite{SchlipfAnnexsteinFrancoSwaminathan95}, 
  Quad  \cite{Dalal96},     UP-Horn   \cite{FourdrinoyGrgoireMazureSais07} and more.
The Related Work   gives a   chronological  account  of such   clausal  
 classes   and their  interrelations.

\vspace{.05cm}
Despite  those  many   clausal classes, so far no attempt  is known to lift the  
Horn     pattern   to the non-clausal level,  
i.e.   no  NC class  is recognized to be the NC   analogous to  the  Horn   class.  
Perhaps, the  Horn  pattern was   considered inherently clausal 
and  not extensible to non-clausal form.  
 %
In this paper, we bridge that gap by  introducing the  hybrid class $\mathbb{H_{NC}}$  of  
{\em Horn-NC formulas,}  which results by suitably amalgamating the Horn-clausal  and the NC formulas,
   or equivalently,  by  {\em appropriately  lifting  the  Horn-clausal pattern to    NC form.}

\vspace{.05cm}
Firstly, the syntactical Horn-NC pattern is established  by  lifting the clausal restriction: {\em a formula  is Horn if all its clauses have (any number of negative 
literals and)  at most one positive literal,}
 to the NC  level, which gives rise to the recursive non-clausal restriction: {\em an NC formula is 
Horn-NC if all its disjunctions have any number of negative disjuncts and 
at most one  non-negative disjunct.} Accordingly, $\mathbb{H_{NC}}$ is the class of  Horn-NC formulas. Note that $\mathbb{H_{NC}}$ naturally subsumes the Horn class.
After such simple definition of $\mathbb{H_{NC}}$, 
 we go deeper into the details and give its 
 detailed definition  compactly and  inductively.
 
 \vspace{0.05cm}
Secondly, we prove the relationships between $\mathbb{H_{NC}}$ and the classes Horn and NC.  
Concretely, we show that $\mathbb{H_{NC}}$ and 
the Horn class are related   in that
(i) $\mathbb{H_{NC}}$ syntactically subsumes the Horn class but both classes are logically equivalent;
and   
 $\mathbb{H_{NC}}$ and the NC class are related  in that (ii)  $\mathbb{H_{NC}}$  contains all NCs
  whose clausal form  is Horn.
     The Venn diagram in {\bf Fig.} 1 below relates    $\mathbb{H}_{\mathbb{NC}}$ to the 
       classes Horn ($\mathbb{H}$), NC ($\mathbb{N_C}$) and Clausal ($\mathbb{C}$).
 
 \vspace{0.05cm}     
       Thirdly,  we provide {\em Non-Clausal  Unit-Resolution}, or UR$_{\mbox{nc}}$,
prove its  completeness for $\mathbb{H_{NC}}$ and  that it allows
to test the satisfiability of 
 $\mathbb{H_{NC}}$ in   polynomial time. This claim makes $\mathbb{H_{NC}}$, so far as we know,
 {\em the first characterized tractable NC class.} 
Such polynomiality allows to easily prove that  {\em   Horn-NC-SAT is P-complete.}

\vspace{-.1cm}   
\begin{center}
 \begin{tikzpicture}
  \begin{scope}[blend group=soft light]
    \fill[black!40!white]   (270:.9) circle (2.3);
    \fill[green!100!white] (230:1.2) circle (1.3);
    \fill[red!100!white]    (310:1.2) circle (1.3);
\end{scope}
  \node at (100:.8)       {\bf \ $\ \mathbb{N_C}$};
  \node at (215:1.6)      {\bf $\mathbb{C}$};
  \node at (325:1.6)    {\bf $\mathbb{H_{NC}}$};
 \node at (270:.9)         {\bf $\mathbb{H}$};
\end{tikzpicture}

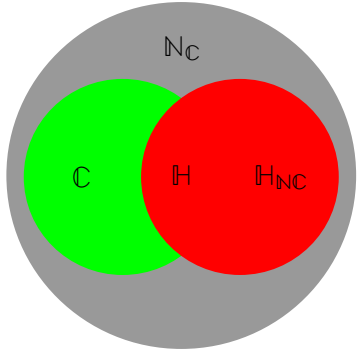
\footnotesize{{\bf Fig. 1.} The Horn$\cdot$NC, Horn, NC  and clausal classes.}
\end{center}

\vspace{-0.2cm} 
Fourthly, we demonstrate the following additional properties of $\mathbb{H_{NC}}$: 
  (i) {\em $\mathbb{H_{NC}}$ is linearly recognizable,}
i.e. deciding whether an  NC  is Horn-NC takes linear time;   
 (ii)  {\em $\mathbb{H_{NC}}$ is more succinct}   \cite{GogicKPS95} than the Horn class, i.e.  
some Horn-NC formulas are exponentially smaller than their equivalent Horn  formulas; and 
(iii) syntactically, $\mathbb{H_{NC}}$ is  {\em exponentially  richer} than  the Horn  class, i.e. for each Horn formula there are exponentially many Horn-NCs.

\vspace{0.05cm} 
We summarize and illustrate our aforementioned contributions through   the  
NC formula  $\varphi  $  below, whose  suffix notation 
will be detailed in Subsection \ref{sec:NCbasis} and  wherein  $ A, B, \ldots $ and
$ \overline{A}, \overline{B}, \ldots $ are positive and negative literals,
respectively,  and $ \phi_1 $,  $ \phi_2 $ and $ \phi_3 $ are   NCs:
\vspace{-.05cm} 
$$\varphi=\{\wedge \ \,  A   \ \, (\vee \ \ \overline{B}  \ \ \{\wedge \ \ (\vee \ \ \overline{D} \ \ \overline{C}   \ \ A) \ \ (\vee \ \ \phi_1 \ \ \{\wedge \ \ \phi_2  \ \  {\overline{A} } \}  ) \ \, B  \} \,)  \ \   \phi_3 \}$$

\vspace{-.05cm} 
\noindent We will show that  $ \varphi $ is Horn-NC when  $ \phi_1 $, $ \phi_2 $
and $ \phi_3 $ are Horn-NC and  at least  one of $ \phi_1 $  or  $ \phi_2 $ is negative. In that case
  we will show that:

\vspace{.15cm}
 $\bullet  $ \ $ \varphi $ can be tested for satisfiability in polynomial time.

\vspace{.03cm}
$\bullet  $ \ $ \varphi $ can be recognized as Horn-NC in  linear time.

\vspace{.03cm}
$\bullet  $ \ $ \varphi $ is logically equivalent to a   Horn formula.

\vspace{.03cm}
$\bullet  $  \ Applying $ \wedge / \vee $ distributivity to $ \varphi $ yields a   Horn formula.

\vspace{.03cm}
$\bullet  $  \ There are exponentially many Horn-NCs equivalent to $ \varphi $. 

\vspace{.03cm}
$\bullet  $ \ $ \varphi $ is exponentially smaller than its equivalent Horn formula.




 \vspace{.15cm} 
In a nutshell,  $\mathbb{H_{NC}}$ gathers  benefits from possessing 
 a salient  expressive power  
and from enjoying appealing properties for efficient reasoning. Hence, 
it is reasonable to expect
that   $\mathbb{H_{NC}}$ in tandem with fast future Horn-NC algorithms 
could be  relevant  
in NC  reasoning to the 
extent to which $\mathbb{H_{NC}}$ could be regarded the NC analogous to
the  Horn clausal class.

\vspace{.05cm}
A direct benefit of this work is that $\mathbb{H_{NC}}$ and UR$_{\mbox{nc}}$ pave the way to develop 
 NC DPLL  reasoners able to  emulate
the efficiency of their clausal counterparts.  We also argue that $\mathbb{H_{NC}}$ and UR$_{\mbox{nc}}$ 
 allow logic programing    and in general, knowledge-based systems: $(i)$ 
    enriching their 
  syntax from simple Horn rules    to Horn-NC rules  
 where   heads and bodies are   NCs with slight  restrictions; and  $(ii)$  
 answering queries with an efficiency  comparable to the clausal efficiency as, both, 
     Horn-NCs can be tested  for  satisfiability polynomially and 
     have only one minimal model (as aforementioned, they are  equivalent to a Horn  formula).
 
\vspace{.05cm}
As a by-product of its properties, $\mathbb{H_{NC}}$  can  potentially
 draw interest from other fields such as Boolean functions and implicational (closure) systems.
 Indeed,  $\mathbb{H_{NC}}$ can serve to analyze  the Boolean functions representable 
by Horn formulas \cite{daglib/0028067}, since
$\mathbb{H_{NC}}$ is  equivalent to the Horn class, is easily recognizable 
and its formulas are   much smaller than their  clausal
representations. Regarding     implicational systems \cite{BertetM10},   
instead of  clausal implications  related to Horn formulas,
general NC implications  
 related to  Horn-NCs can be  envisioned. 
  Further   applications using the $\mathbb{H_{NC}}$ properties
 are given in the related work.

\vspace{.05cm}
The Future Work section outlines a substantial 
number of   research directions which pivot on two research axes: 
(i)      $\mathbb{H_{NC}}$ and  UR$_{\mbox{nc}}$   can be smoothly extended  
 beyond propositional logic to richer logics and    adapted to different
reasoning problems; and (ii) our  work can be used as a lever  to develop the  NC paradigm;
for example, our   definition of  UR$_{\mbox{nc}}$ is  the basis to conceive a 
new definition of Non-Clausal Resolution.

\vspace{.05cm} 
  This  paper continues as  follows.
  Section \ref{sec:NCbasis}  presents     background and terminology.   
   Section \ref{sec:definition}    defines  
     $\mathbb{H_{NC}}$ and relates $\mathbb{H_{NC}}$ 
     to the classes NC  and Horn. Section  \ref{sec:polynomial}
    introduces  UR$_{\mbox{nc}}$  and proves the tractability of $\mathbb{H_{NC}}$.   Section
   \ref{sec:semanticalsyntactical} 
    demonstrates further  properties of $\mathbb{H_{NC}}$. 
    Section \ref{sec:relatedwork} and \ref{sec:futurework} focus on
  related work   and  future work, respectively. The last section   
 summarizes the main contributions.

\section{Background,  Notation and Terminology} \label{sec:NCbasis}

  This section     presents  background  on   propositional   non-clausal logic     
  (the reader is referred  to e.g.  \cite{DBLP:books/daglib/0029942} for  a more complete  background)
   and introduces our notation and  terminology.



\begin{definition} \label{def:alphCNF}
 Positive  literals are noted $\{A,   B, \ldots\}$   and   negative 
ones $\{\overline{A},  \overline{B}, \ldots\}$.  $\mathbb{L}$ is the set of literals.
    A  clause    is a
 disjunction of  literals.    A    clause  with    at most  
  one positive literal is Horn. A conjunction
of    clauses     is a  clausal     formula.  A conjunction  of Horn clauses  is a Horn formula. 
  $\mathbb{C}$ and $\mathbb{H}$ are the sets of  clausal and Horn formulas, respectively. 
\end{definition}


\begin{definition}  The   NC  alphabet  is formed by the sets of:  
  constants  \{{\bf T,\,F}\}, literals $\mathbb{L}$, connectives  \{$\neg,\vee,  \wedge\}$
  and  auxiliary symbols:   (,  ),  \{   and    \}. 
\end{definition}

\noindent  For the sake of {\em readability of NC formulas,} we   employ: (1) 
 the prefix notation       
     because  it   requires only one connective  $\vee$ or $\wedge$
      per formula,      
       while infix notation requires
   $k-1$ ($k$  is  the arity of  $\vee$ and $\wedge$); and
 (2) two   formula    delimiters,  $\{\wedge  \,\ldots \,\}$ for conjunctions
and    $(\vee  \,\ldots \,)$ for disjunctions 
(Definition \ref{def:NNFformulas}), 
to  distinguish  them  inside  nested   NCs.
%
%
The    
 negation normal form  formulas (NNFs) do not use  
 the connective  $\neg$. We give next our  notation for  NNFs.

\begin{definition}  \label{def:NNFformulas} The  set  $\mathbb{N_{NF}}$ of
NNF formulas is   defined in the usual  way:

\begin{itemize}
\item   $\{{\bf T,\,F}\} \cup \mathbb{L} \,\subset \,\mathbb{N_{NF}}$.

\vspace{-.15cm}
\item  If  for all $i \in \{1, \ldots k \}$,  $\varphi_i \in \mathbb{N_{NF}}$  \,then    
\,$\{\wedge  \ \varphi_1 \ldots \varphi_{i}  \ldots    \varphi_k\} \in  \mathbb{N_{NF}}$.

\vspace{-.15cm}
\item   If for all $i \in \{1, \ldots k \}$, $\varphi_i \in \mathbb{N_{NF}}$ \,then  \,$\,(\vee  \ \varphi_1  \ldots 
\varphi_{i} \ldots   \varphi_k) \in  \mathbb{N_{NF}}$.
\end{itemize}


\vspace{-.15cm}
 -- \ $\{\wedge    \,\varphi_1 \ldots  \varphi_{i}  \ldots  \varphi_k \}$ and any $\varphi_i$ 
 are called    conjunction and conjunct,  respectively.

\vspace{.1cm}
  -- \ $(\vee   \,\varphi_1  \ldots  \varphi_{i} \ldots  \varphi_k )$ and any $\varphi_i$ are called  
disjunction  and disjunct, respectively. 

\vspace{.1cm}
-- \ $\langle \odot  \,\varphi_1  \ldots  \varphi_{i} \ldots  \varphi_k\rangle$    stands for both 
 $(\vee   \,\varphi_1  \ldots  \varphi_{i} \ldots   \varphi_k ) \mbox{\ and \ }
\{\wedge   \,\varphi_1  \ldots  \varphi_{i}  \ldots  \varphi_k \}.$  
\end{definition}

\begin{example} \label{ex:NNF-NC} $\varphi_1$ and  $\varphi_3$ below are NNF while
      $\varphi_2$ and  $\varphi_4$ are not. 
\begin{itemize}
\item [--]    $\varphi_1=(\vee \ \ \{\wedge \ \ \overline{A} \ \   {\bf T} \,\} \  
\  \{\wedge  \  \   (\vee   \ \   \overline{A} \  \ C \, ) \ \ {\bf T} \ \ \{\wedge \ 
\ D \ \   (\vee \  \ A  \ \ \overline{B} \,) \,\} \,\}  \, )$.  

\vspace{-.15cm}
\item  [--]   $\varphi_2=\{\wedge    \ \ D \  \
\neg (\vee  \ \ \{\wedge \  \ \overline{A} \  \ B \,\}  \ \ {\bf F}
 \ \ (\vee \ \ A \  \   C \, ) \, ) \, \}.$
 
 \vspace{-.15cm}
\item [--]  $\varphi_3= \{\wedge \ \ \varphi_1 \ \ (\vee  \ \ \{\wedge  \ \ A \ \ \overline{C}\} 
\  \ \overline{B})
\ \ (\vee \ \  \varphi_1  \  \ {\bf F} \ \ A) \,\}.$

\vspace{-.15cm}
\item [--]  $\varphi_4 = (\vee \ \ \varphi_2 \ \ \{\wedge  \ \ A  \ \ 
(\vee \ \ \varphi_1 \ \ \overline{D} \ \   \varphi_3) \,\} \ \ 
\{\wedge \ \ \varphi_2 \ \ B \ \ (\vee \ \   \varphi_1 \ \ \varphi_3) \,\} \,)$ \qed
 
\end{itemize} 

\end{example}

\begin{definition}  \label{def:NCformulas}
    The set $\mathbb{N_C}$ of NC formulas   is   built  inductively as     $\mathbb{N_{NF}}$ 
    but over  the whole NC alphabet, namely including also the connective $ \neg $ (we omit the formal details).
\end{definition}

\begin{example}   $\varphi_2$  and $\varphi_4$  in   Example \ref{ex:NNF-NC}, 
which were  not in $\mathbb{N_{NF}}$,  are  in $\mathbb{N_C}$.
\end{example}


\begin{definition} \label{def:sub-for} Sub-formulas are inductively defined as 
follows. The unique sub-formula of an atom ($\{{\bf T,\,F}\} \cup \mathbb{L}$) is the atom  itself. 
The sub-formulas of $\varphi$ are $\varphi$ itself more  the sub-formulas of either (i)      $\varphi'$, if $\varphi=\neg \varphi'$,  
or (ii)    the     $\varphi_i$'s, if  $\varphi=\langle \odot \  \varphi_1 \ldots \varphi_i \ldots \varphi_k \rangle$.
\end{definition}

\begin{definition} \label{def:graphformula} 
NCs  are   representable by trees if: (i)
each atom  is a   {\em   leaf node} and
each  occurrence of   
  connectives  is   an {\em  internal node;} and
(ii)   the arcs are given by:  (a)
      each sub-formula   $\neg \varphi$  is   an arc
   linking  the nodes  of  $\neg$ and    
   of the connective of   $\varphi$; and (b) 
  each sub-formula  
$\langle \odot \  \varphi_1 \ldots  \varphi_{i}  \ldots \,\varphi_k \rangle$    
 is  a  $k$-ary hyper-arc
 linking  the nodes of  $\odot$  and     
 the  nodes of, for every $i$,  $\varphi_i$  if $\varphi_i$ 
 is an  atom and of its connective  otherwise.
\end{definition}

\begin{example}  The   tree    representing \,$\varphi_2$ in Example \ref{ex:NNF-NC} is given in Fig. 2.
\end{example}
\begin{center}
\begin{tikzpicture}[sibling distance=10em,
  every node/.style = {shape=rectangle, rounded corners,
    draw, align=center,
    top color=pink, bottom color=pink!100}]]
  \node {$\wedge$}
    child { node {$\overline{C}$} }
    child { node {$\neg$}
      child { node {$\vee$} 
        child { node {$\wedge$} 
           child { node {$\overline{A}$} }
           child { node {$B$}    } }
        child { node {\bf F} } 
        child  { node {$\vee$} 
           child { node {$\overline{A}$}  }  
           child { node {$\overline{C}$}  }}}};
\end{tikzpicture}

\vspace{-.1cm}
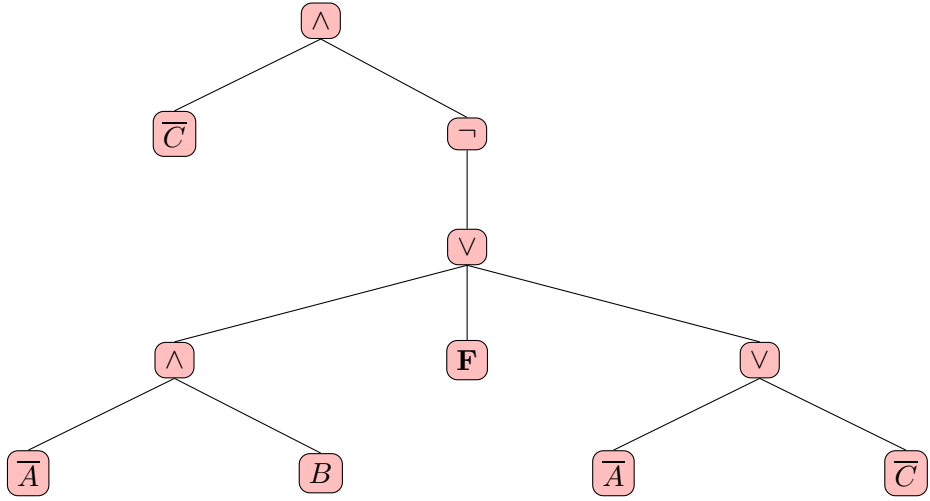
{\footnotesize{\bf Fig. 2.}   Tree of $\varphi_2$.}
\end{center}
\vspace{-.7cm}
\qed

\vspace{.cm}
\noindent There exist different,  bi-dimensional   graphical models   \cite{Bibel81,JainBartzisClarke06}  
Our approach also applies when   NCs are modeled and implemented
as   directed acyclic graphs (DAGs), 
which are  more  general   than trees and
 allow for important savings in both  space and  time.

\begin{definition} \label{def:DAGs}  An  NC formula $ \varphi $ is modeled
by a DAG when each  sub-formula $ \phi $  
 is modeled 
by a unique DAG $D_\phi$ and   each $ \phi $-occurrence by a  pointer 
to (the root of) $D_\phi$.
\end{definition}

\begin{example} 
Each sub-formula $ \varphi_1 $ to $ \varphi_3 $ 
in $ \varphi_4 $ from Example \ref{ex:NNF-NC} requires one 
DAG and two  sub-formula pointers.
\end{example}



\noindent An {\bf interpretation} 
  $\mathcal{I}$
maps  $\mathbb{N_C}$  onto  $\{0,1\}$  and maps constants as follows:
$\mathcal{I}({\bf F})=\mathcal{I}(\,(\vee)\,)=0$ \ and \ 
$\mathcal{I}({\bf T})=\mathcal{I}(\,\{\wedge\}\,)=1.$ As the interpretation
of negated, conjunctive and disjunctive formulas is well-known, 
by space reasons we omit their formal definitions.

\begin{definition}
 An interpretation    $\mathcal{I}$ is a   model of   \,$\varphi$  if  \,$\mathcal{I}(\varphi)=1.$ 
 If $\varphi$  has a model then it is    satisfiable and otherwise unsatisfiable.
  $\varphi$ and $\varphi'$ are    equivalent, noted
 $\varphi \equiv \varphi'$,   if   
  \ $\forall \mathcal{I}$, 
  $\mathcal{I}(\varphi)=\mathcal{I}(\varphi')$.
 $\varphi'$ is  logical consequence of $\varphi$, noted
 $\varphi \models \varphi'$,   if   
  \ $\forall \mathcal{I}$,   $\mathcal{I}(\varphi) \leq \mathcal{I}(\varphi')$.
\end{definition}

\begin{definition}  We   define: SAT, Horn-SAT,
NC-SAT and Horn-NC-SAT are the problems of testing the satisfiability of clausal,
Horn, NC and Horn-NC formulas, respectively.
%
%
%
%
\end{definition}

\noindent {\bf Complexity.}  SAT \cite{Cook71} and NC-SAT are   NP-complete,  Horn-SAT is linear   
 \cite{DowlingGallier84,ItaiMakowsky87,Minoux88, Scutella90,DuboisAndreBoufkhadCarlier96}.
Here,  
 the Horn-NC class is defined and proved that  Horn-NC-SAT is polynomial. 
 

\begin{definition} \label{def:simpl}    
 Constant-free,  equivalent      formulas   are  straightforwardly obtained 
  by applying to  sub-formulas the  next  simplifying  rules:  

\vspace{.2cm}
$\bullet$ Replace: $\{\wedge \  {\bf F} \  \varphi \,\} \leftarrow  
{\bf F}$;     \ \ 
  $(\vee \  {\bf T} \  \varphi \,) \leftarrow {\bf T}$; \ \
  $\{\wedge \  {\bf T} \  \varphi \,\} \leftarrow \varphi$;  
  \ \ $(\vee \  \,{\bf F} \  \varphi \,)$     $\leftarrow$  $\varphi$
\end{definition}

\noindent {\bf Remark.}  For simplicity and since constant-free,    equivalent  formulas are 
 easily obtained,  hereafter we will consider only constant-free formulas.

\section{The Horn-NNF and Horn-NC Classes} \label{sec:definition}
 We give first the   definition of $\mathbb{H_{NF}}$
and then that of  $\mathbb{H_{NC}}$. 
  We will use the following abbreviations: 
 \  HNF for   Horn-NNF \  and \  HNC for  Horn-NC.
%



\subsection{The Horn-NNF formulas:  $\mathbb{H_{NF}}$}


 \begin{definition}  \label{def:negative} Negative  formulas
 are   NNF formulas having solely negative literals.   $\mathbb{{N_G}}$ is the set of
  negative NNF formulas. 
\end{definition}


\begin{example} An example of negative formula is $(\vee \ \ \{\wedge   \  \,\overline{A} \  \,\overline{C} \, \} 
\ \ \{\wedge \ \,\overline{E} \  \,(\vee \ \,\overline{A}  \ \,\overline{B} \,)\,\}\,)$. \qed
\end{example}

  
Next we  lift the   the Horn pattern
 to the non-clausal level as follows:

\begin{definition} \label{theorem:visual} An NNF   is HNF all its disjunctions have 
any number of negative disjuncts and at most one
 non-negative disjunct.  
We denote $ \mathbb{H_{NF}} $   the class of HNF formulas.
\end{definition}

Clearly Horn formulas are HNF, namely $ \mathbb{H} \subset \mathbb{H_{NF}} $.

\vspace{.15cm}
\noindent {\bf  Remark}.    As  Definition   \ref{theorem:visual} is not concerned with how 
NNFs are represented,  
     our approach  also applies when   NNFs  are  represented 
     by  DAGs and not just by trees. Yet, for the sake of simplicity, we will use  NNFs
representable by   trees  throughout this article.

\begin{example} \label{exam:simple} Let us consider $\varphi_1$ and $\varphi_2$ below.
As  $\varphi_1$ has two disjuncts and
  only one is non-negative,
  $\varphi_1$ is HNF. Yet $\varphi_2$ has two non-negative disjuncts and so 
it is not HNF. 
 \begin{itemize}
\item   $\varphi_1 = (\vee  \   \  \{\wedge \  \ \overline{B} \ \  \overline{D}\} \  \ 
\{\wedge \  \ C \   \ A \,\} \,) \qquad \varphi_2 = (\vee   \   \   \{\wedge  \ \ \overline{B} \   \ D\} \  \ 
\{\wedge \  \ C \  \ \overline{A} \,\} \,).$ 
 \qed
\end{itemize} 
\end{example}

 \begin{example} \label{Ex:morecomplex} Below  we consider     
  $\varphi$   and $\varphi'$   
    resulting from $\varphi$ by 
switching  its  left-most $\overline{A}$  for A. All the   disjunctions of $\varphi$, i.e. $(\vee   \    \overline{A} \   C)$, 
$(\vee   \   A \   \overline{B})$ and   $\varphi$ itself, 
have exactly one non-
 \begin{itemize}
 \item $\varphi=(\vee \quad \overline{A} \ \ \{\wedge \  \   
(\vee   \ \   \overline{A} \  \ C \,) 
\quad \{\wedge \ \ D \ \   (\vee \ \ A  \ \ \overline{B} \,) \,\}  \,\}  \, )$.

\item  $\varphi'=(\vee \quad A \ \ \{\wedge \  \   
(\vee   \ \   \overline{A} \  \ C \,) 
\quad \{\wedge \ \ D \ \   (\vee \ \ A  \ \ \overline{B} \,) \,\}  \,\}  \, )$.

 \end{itemize}  
   
\noindent  
negative disjunct;  so $\varphi$  is HNF. Yet, $\varphi'=(\vee \  A \ \phi)$,
 $\phi$  being  non-negative;  so, as  $\varphi'$ has two
non-negative disjuncts,   $\varphi'$ is not HNF. \qed
\end{example}

 \begin{proposition} \label{propos:subformulasHNF} All sub-formulas of an HNF formula are HNF.
 \end{proposition}
 
 The proof follows immediately from Definition \ref{theorem:visual}.
%
Note that the converse   does not hold since there are non-HNF formulas
all of whose  sub-formulas are HHF.

\vspace{.1cm}
 Towards  a   fine-grained definition of $\mathbb{{H}_{NF}}$,  first we   inductively specify
        $\mathrm{{HNF}}$ conjunctions (Lemma \ref{def:HNNFconjunc}) and 
   $\mathrm{{HNF}}$ disjunctions (Lemma \ref{def:disjunHNNF}),  
 and then   we {\em compactly}  specify   $\mathbb{{H}_{NF}}$   
 by  embedding  both specifications    into an inductive function  
 (Definition \ref{cor:syntacticalNNF}).

\vspace{.1cm}
Conjunctions of Horn  clausal  formulas
      are    Horn too, 
 and   a similar kind of {\em Horn-like compliance} 
 also holds in NC form,   viz.  
    conjunctions of HNF formulas are     HNF too.
  
  \begin{lemma} \label{def:HNNFconjunc} 
  Conjunctions of   
      $\mathrm{{HNF}}$ formulas  are    $\mathrm{{HNF}}$ as well, formally
  $$\{\wedge \  \varphi_1   \,\ldots\,  \varphi_{i} \ldots \,\varphi_k\} \in \mathbb{{H}_{NF}}
  \mbox{\em \ \ iff \ \ for }   1 \leq i \leq k, \ \varphi_i  \in \mathbb{{H}_{NF}}.$$
  \end{lemma}
  
The proof is straightforward.

\begin{example}  \label{ex:Semant-Conjunt} If  $\mathrm{H}_1$ is  Horn 
and $\varphi \in \mathbb{{H}_{NF}}$,  then   $\{\wedge \ \mathrm{H}_1  \ (\vee \ \overline{D} \ A \,) \ \varphi\} \in \mathbb{{H}_{NF}}$
\qed
\end{example}

   It is not hard to check  that the definition of HNF disjunction  in 
   Definition \ref{theorem:visual} can be 
  equivalently   expressed in the next recursive manner: 
 {\em an  NNF disjunction is HNF if it has any number of negative disjuncts and 
 one HNF disjunct,}
 which leads  to:

\begin{lemma} \label{def:disjunHNNF}  
A disjunctive NNF  
$\varphi=(\vee \ \varphi_1 \ldots \varphi_i \ldots  \varphi_k)$ with  $k \geq 1$ disjuncts 
belongs to $ \mathbb{{H}_{NF}} $  iff  
 it has one $\mathrm{{HNF}}$ and $k-1$  negative   disjuncts,    
 formally    
$$(\vee \ \varphi_1 \ldots \varphi_i \ldots     \varphi_k) \in \mathbb{{H}_{NF}} \mbox{\em \ \ iff  there is} \ i \ \mbox{\em s.t.} \ \varphi_i \in \mathbb{{H}_{NF}} 
\  \mbox{\em and} \ \mbox{\em for all }  j \neq i,  \varphi_j \in \mathbb{{N_G}}.$$ 
\end{lemma}

\begin{niceproof} {\bf If.} Since   the  sub-formulas 
$\forall j, j \neq i, \varphi_j $ 
have no positive literals,    the non-negative disjunctions
of $\varphi= (\vee \ \varphi_1 \ldots   \varphi_i \ldots     \varphi_k) $ are those 
  in $ \varphi_i $ plus $ \varphi_i $ and $ \varphi $ themselves.
Given that  by hypothesis $ \varphi_i \in \mathbb{H}_\mathbb{NF}  $  and that
$\forall j, j \neq i, \varphi_j $ 
have no positive literals, then  $\varphi=(\vee \ \varphi_1 \ldots   \varphi_i \ldots     \varphi_k)$ 
satisfies Definition \ref{theorem:visual}, and so $\varphi \in \mathbb{H}_\mathbb{NF}$.
 {\bf Iff.} It is done 
    by contradiction: if (i) $ \varphi_i \notin \mathbb{H}_\mathbb{NF} $  or (ii)
 $\exists i,j, i \neq j,  \varphi_i,\varphi_j \notin \mathbb{{N_G}}$,  then 
$  (\vee \ \varphi_1 \ldots \varphi_i \ldots     \varphi_k) \notin \mathbb{H}_\mathbb{NF}$. 
A similar proof is given in Theorem \ref{th:kdisjuncts} and so, by space reasons, we omit the details.
\end{niceproof}


From Lemma \ref{def:disjunHNNF}, we have:
(A) non-recursive $\mathrm{{HNF}}$ disjunctions are   Horn clauses; 
(B) NNF disjunctions with all negative disjuncts are $\mathrm{{HNF}}$; and
(C)  NNF disjunctions with $ k \geq 2$      non-negative disjuncts are not 
$\mathrm{{HNF}}$.  
 %
 Next, we   reexamine 
Examples  \ref{exam:simple} and \ref{Ex:morecomplex} but this time bearing Lemma \ref{def:disjunHNNF} in mind.

\begin{example} \label{exam:NF} Below we analyze   $\varphi_1$  and $\varphi_2$  from  Example \ref{exam:simple}.
\begin{itemize}

\item    $\varphi_1 = (\vee  \   \{\wedge \   \overline{B} \   \overline{D}\} \   
\{\wedge \   C \    A \,\} \,).$  

-- Clearly
$\{\wedge \   \overline{B} \   \overline{D}\} \in \mathbb{{N_G}}$  and  by Lemma \ref{def:HNNFconjunc},
$\{\wedge \   C \    A \,\} \in \mathbb{{H}_{NF}}$.

--  According to Lemma \ref{def:disjunHNNF}, 
  $\varphi_1 \in \mathbb{{H}_{NF}}$.

\item   $\varphi_2 = (\vee   \   \{\wedge  \  \overline{B} \    D\} \  \ 
\{\wedge \   C \   \overline{A} \,\} \,)$.

-- Clearly $\{\wedge  \  \overline{B} \  D\} \notin \mathbb{N_G}$ 
and  $\{\wedge \ C \ \overline{A} \,\} \notin \mathbb{{N_G}}$;
\ by Lemma \ref{def:disjunHNNF}, 
  $\varphi_2 \notin \mathbb{{H}_{NF}}$. \qed

\end{itemize}   
\end{example}

\begin{example} \label{ex:disjunHNNF} Let us consider again     $\varphi$  and $\varphi'$ in Example \ref{Ex:morecomplex}  
     recalling that $\varphi'$ results from $\varphi$ by  
  just  switching  its left-most  $\overline{A}$ for $A$.   
\begin{itemize}
\item By  Lemma \ref{def:disjunHNNF}, $(\vee \ \overline{A} \   C) \in \mathbb{{H}_{NF}}$ 
and $(\vee \  A  \  \overline{B})  \in \mathbb{{H}_{NF}}$.

\item  By  Lemma  \ref{def:HNNFconjunc},  $ \{\wedge \ \ D \    \,(\vee \  A  \  \overline{B})  \}\in \mathbb{{H}_{NF}}$. 
 
\item  By  Lemma  \ref{def:HNNFconjunc},   $\phi=\{\wedge \  (\vee  \  \overline{A} \   C \,) \ 
 \{\wedge \ \ D \    \,(\vee \  A  \  \overline{B})  \} \} \in \mathbb{{H}_{NF}}$.
 

\item We have: $\varphi= (\vee \ \overline{A} \ \,\phi \,)$. 
\ \ Since $\overline{A} \in \mathbb{{N_G}}$
and   $\phi \in \mathbb{{H}_{NF}}$, by Lemma \ref{def:disjunHNNF}, $\varphi \in \mathbb{{H}_{NF}}$.
    
\item We have:  
$\varphi'=(\vee \   A \ \phi \,)$. \ \ 
Since $A,\phi \notin \mathbb{{N_G}}$,   by Lemma \ref{def:disjunHNNF},  $\varphi' \notin \mathbb{{H}_{NF}}$. \qed

\end{itemize}

\end{example}

\noindent  By merging    Lemmas  \ref{def:HNNFconjunc}  and    \ref{def:disjunHNNF},    $\mathbb{{H}_{NF}}$ is  compactly and inductively defined as follows.

\begin{definition} \label{cor:syntacticalNNF}   
We inductively define the set   of formulas    $\mathbb{\overline{H}_{NF}}$ exclusively from 
    the  rules below,  wherein
 $k \geq 1$ and $\mathbb{L}$ is the set of   literals.
\begin{itemize}

\item [(1)]  $\mathbb{L} \subset \mathbb{\overline{H}_{NF}}.$                    \hspace{10.05cm}                 

\item [(2)] $\mathrm{If} \ \ \forall i, \,\varphi_i  \in \mathbb{\overline{H}_{NF}} \  \ \mathrm{then} \  
  \ \{\wedge \  \varphi_1   \,\ldots\, \varphi_{i} \ldots  \varphi_k\} \in \mathbb{\overline{H}_{NF}}.$ \hspace{3.08cm} 

\item [(3)]   $\mathrm{If} \ \ \varphi_i \in \mathbb{\overline{H}_{NF}} \ \, \mathrm{and} \ \,
 \forall j \neq i$,  $\varphi_j \in \mathbb{{N_G}} \ \, \mathrm{then} \ \, (\vee \ \varphi_1 \ldots \varphi_i \ldots   \,\varphi_k) 
 \in \mathbb{\overline{H}_{NF}}.$ \hspace{.11cm} 
 
\end{itemize}

\end{definition}


We prove next that  $ \mathbb{\overline{H}_{NF}} $ coincides with $\mathbb{{H}_{NF}}$, that is,
Definition \ref{cor:syntacticalNNF}  corresponds to the recursive and compact definition of $\mathbb{{H}_{NF}}$.
Besides,  the specification  
in  Definition  \ref{cor:syntacticalNNF} yields an optimal strategy to decide whether an NNF   
 is $\mathrm{{HNF}}$, which is   proven in Subsection  \ref{subsec_SynProp}. 
In fact,    a recognition algorithm is designed   and
 its {\em linear behavior} proved.

\begin{theorem} \label{th:HNFequality}  We have: \ $\mathbb{\overline{H}_{NF}}=\mathbb{{H}_{NF}}$.

\end{theorem}

\begin{niceproof}  We prove first $\mathbb{\overline{H}_{NF}} \subseteq \mathbb{{H}_{NF}}$ and then $\mathbb{\overline{H}_{NF}} \supseteq \mathbb{{H}_{NF}}$. The first relation $\mathbb{\overline{H}_{NF}} \subseteq \mathbb{{H}_{NF}}$
    is  proven by structural induction, where 
    ({1}) $\mathbb{L} \subset \mathbb{{H}_{NF}}$   holds trivially, as follows.
    
\vspace{.05cm} 
 ({2}) The non-recursive $\mathbb{\overline{H}_{NF}}$ conjunctions are literal  conjunctions, which trivially verify
 Definition  \ref{theorem:visual} and so are in $\mathbb{{H}_{NF}}$. Further, assuming
 that for an induction step $\mathbb{\overline{H}_{NF}} \subseteq \mathbb{{H}_{NF}}$ holds
 and that    $\varphi_{i} \in \mathbb{\overline{H}_{NF}}, 1 \leq i \leq k$, 
 then any formula $\{\wedge \  \varphi_1   \,\ldots\, \varphi_{i} \ldots  \varphi_k\}$  added 
 in (2) to $\mathbb{\overline{H}_{NF}}$ belongs also to $ \mathbb{{H}_{NF}}  $   by Lemma \ref{def:HNNFconjunc}. So
  $\mathbb{\overline{H}_{NF}} \subseteq \mathbb{{H}_{NF}}$  holds. 

\vspace{.05cm}
({3}) Assuming that for a given recursive level  $\mathbb{\overline{H}_{NF}} \subseteq \mathbb{{H}_{NF}}$ holds,
in the next recursion, only  disjunctions $ \varphi $  in ({3}) are added to 
$\mathbb{\overline{H}_{NF}}$. 
But  the  condition of ({3})  and that of 
 Lemma \ref{def:disjunHNNF} are equal; so by  Lemma \ref{def:disjunHNNF},  
 $ \varphi $  is in  $\mathbb{{H}_{NF}}$ too.  Therefore $\mathbb{\overline{H}_{NF}} \subseteq \mathbb{{H}_{NF}}$ holds.
 
 \vspace{.05cm} 
$\bullet$ $\mathbb{{H}_{NF}} \subseteq \mathbb{\overline{H}_{NF}}$. 
Given  that the structures to define $\mathbb{{N}_{NF}}$ and $\mathbb{\overline{H}_{NF}}$ 
in Definition \ref{def:NNFformulas}  and Definition \ref{cor:syntacticalNNF}, respectively,    
   are equal, the potential inclusion  of each  NNF  
   formula $ \varphi$ in   $\mathbb{\overline{H}_{NF}}$ is systematically considered. 
 Further, the statement:
  if $ \varphi \in  \mathbb{{H}_{NF}}$ then $ \varphi \in  \mathbb{\overline{H}_{NF}}$, 
    is    proven by structural induction on the depth of formulas 
    and by using a  reasoning   similar to that of 
  the previous $\mathbb{\overline{H}_{NF}} \subseteq \mathbb{{H}_{NF}}$ case and also by using  Lemmas \ref{def:HNNFconjunc}
  and  \ref{def:disjunHNNF}.
\end{niceproof}

\begin{example} \label{ex:ExamComplet} We regard again   $\varphi$ and $\varphi'$ from Example \ref{ex:disjunHNNF}: 
\begin{itemize}
\item By   ({3}), \  $(\vee \ \overline{A} \   C) \in \mathbb{{H}_{NF}}$.  \qquad \qquad \ \
$(\vee \  A  \  \overline{B}) \in \mathbb{{H}_{NF}}$. 

\item  By   ({2}), \ $\{\wedge \ \ D \    \,(\vee \  A  \  \overline{B})  \} 
\in \mathbb{{H}_{NF}}$. \quad $\phi=\{\wedge \  (\vee  \  \overline{A} \   C \,) \  \{\wedge \ \ D \    \,(\vee \  A  \  \overline{B})  \} \} \in \mathbb{{H}_{NF}}$.

\item By   ({3}), \ $\varphi= (\vee \ \overline{A} \ \,\phi \,) \in \mathbb{{H}_{NF}}$. \quad \quad \ \ $\varphi'=(\vee \   A \ \phi \,) \notin \mathbb{{H}_{NF}}$. \qed


\end{itemize}
\end{example}


\begin{example} We now analyze a more complete NNF, concretely 
  the one from the Introduction. Let us consider $ \varphi $ below  
wherein $ \phi_1, \phi_2$ and $\phi_3 $ are NCs:
$$\varphi=\{\wedge \ \,  A   \ \, (\vee \ \ \overline{B}  \ \ \{\wedge \ \ (\vee \ \ \overline{D} \ \ \overline{C}   \ \ A) \ \ (\vee \ \ \phi_1 \ \ \{\wedge \ \ \phi_2  \ \  {\overline{A} } \}  ) \ \, B  \} \,)  \ \   \phi_3 \}$$
We check under which  conditions verified by $ \phi_1,  \phi_2 $ and $ \phi_3 $, 
 $ \varphi $ is indeed HNF. The disjunctions of $ \varphi $ and the proper $ \varphi $ 
 can be rewritten as follows: 
\begin{itemize}

\item  $ \omega_1= (\vee \ \ \overline{D} \ \ \overline{C}   \ \ A) $. \qquad \quad \qquad \qquad \ $\omega_2= (\vee \ \ \phi_1 \ \ \{\wedge \ \ \phi_2  \ \  {\overline{A} } \}  ) $.

\vspace{-.15cm}
\item  $ \omega_3=(\vee \ \ \overline{B} 
 \ \ \{\wedge \ \ \omega_1 \ \ \omega_2 \ \, B  \} \,).$
 \qquad \quad $ \varphi=\{\wedge \   A   \    \omega_3 \ \phi_3 \} .$

\end{itemize}
 
\noindent We analyze one-by-one such disjunctions and finally the proper $ \varphi $: 
\begin{itemize}

\item   $ \omega_1$: Trivially, $ \omega_1 $ is Horn. 

\vspace{-.15cm}
\item     $ \omega_2 $:  
$ \omega_2 $ is HNF  \ if \  $ \phi_1, \phi_2 \in \mathbb{H_{NF}}$ \ and if 
at least one of $ \phi_1 $ or $ \phi_2$  is negative.

\vspace{-.15cm}
\item    $ \omega_3 $:  $ \omega_3$ is  HNF \ if \   $\omega_2 \in \mathbb{H_{NF}}$  (as $\omega_1 \in \mathbb{H_{NF}}$).
 
 \vspace{-.15cm}
\item     $ \varphi $: \ 
 $ \varphi$  is HNF  \ if \ $ \omega_2 \in \mathbb{H_{NF}}$ (see previous line) and $  \phi_3 \in \mathbb{H_{NF}}$.
\end{itemize}
 
\noindent Recapitulating,   the second ($ \omega_2 $) and fourth  ($ \varphi $)
conditions entail that $ \varphi $ is HNF only if
$ \phi_1 $, $ \phi_2 $ and $ \phi_3 $ are HNF and if at least one of $ \phi_1 $ or $ \phi_2 $ is negative.
Since the first condition is subsumed by Proposition \ref{propos:subformulasHNF},  we can conclude that $ \varphi $ is HNF only if its sub-formulas are HNF  and if at least one of $ \phi_1 $ or $ \phi_2 $ is negative. \qed
\end{example}

\subsection{Relating  $\mathbb{H_{NF}}$ to the  Horn and NNF Classes} \label{sec:syntaxNC}

We prove  that $ \mathbb{{H}_{NF}}$ 
and   the  Horn class are semantically equivalent and    specify the  NNF  fragment
that forms $ \mathbb{{H}_{NF}}$. 
To facilitate the reading of the section, 
 the  proofs
 of the theorems  are relegated to Subsection \ref{sect:HNCH*NC}.
%
%
%
    A new simple concept  is introduced next.  

\begin{definition} \label{def:ECNF}  For every $\varphi \in \mathbb{N_{NF}}$,  we define   $cl(\varphi)$ as
    the unique clausal formula    that
  results from   applying  
 $\vee/\wedge$  distributivity to $\varphi$ until a clausal formula, viz. $cl(\varphi)$, is obtained.
 We will  call $cl(\varphi)$ the clausal form of $\varphi$.
\end{definition}

\begin{example} \label{exa:ECNF} 
Applying $\vee/\wedge$ distributivity to $\varphi_1$  in  Example \ref{exam:simple},  
 one obtains:

\vspace{.25cm}
\hspace{1.cm}$cl(\varphi_1)=\{\wedge \  \ (\vee \ \overline{B}  \, \ \,C) \  \    
 (\vee \ \,\overline{B} \ A) \ \ (\vee \ \overline{D} \ \,C\,) \ \ 
 (\vee \ \ \overline{D} \ \,A )\,\}.$ \qed
 

\end{example}


  \begin{proposition} \label{cor:firstcor} We have \ $\varphi \equiv cl(\varphi)$.
  \end{proposition}
  
   
  The proof is trivial. We next show, in Theorem  \ref{theorem:first}, Corollary \ref{cor:firstcorprimer} 
  and Theorem \ref{prop:sem-NNF}, that  $cl(\varphi)$ allows 
  to  relate    $\mathbb{{H}_{NF}}$ to the classes Horn and  NNF.

\begin{theorem}  \label{theorem:first} 
The  clausal form of all HNF formulas is Horn, i.e. 
 $\forall \varphi \in \mathbb{H_{NF}}:  cl(\varphi) \in \mathbb{H}.$
\end{theorem}

\begin{niceproof} See Subsection \ref{sect:HNCH*NC}.
\end{niceproof}

  Theorem \ref{theorem:first} and Proposition \ref{cor:firstcor}   yield the next  semantical characterization of $ \mathbb{H_{NF}}$.

\begin{corollary} \label{cor:firstcorprimer}
The classes  $ \mathbb{H_{NF}}$ and $ \mathbb{H}$ are semantically equivalent: each formula in a class
  is equivalent to another formula in the other class.
\end{corollary}

\begin{niceproof} On the one hand, by Proposition \ref{cor:firstcor} and Theorem \ref{theorem:first}, for 
 every HNF formula $ \varphi$,  we have  $\forall \varphi \in \mathbb{H_{NF}}: \varphi \equiv cl(\varphi) \in   \mathbb{H}$. 
On the other hand, 
   $ \mathbb{H} \subset \mathbb{H_{NF}} $.
\end{niceproof}


 Corollary \ref{cor:firstcorprimer}  entails that $ \mathbb{H_{NF}}$ can represent
the same Boolean functions   that the Horn class, viz. 
 the Horn functions \cite{daglib/0028067} (see Section \ref{sec:Applcac}).
 %
  The next theorem specifies which NNF formulas are included in  $\mathbb{{H}_{NF}}$.

\begin{theorem} \label{prop:sem-NNF} 
All  NNF formulas  $ \varphi $ whose clausal form is Horn are HNF, namely:
$$ \forall \varphi \in \mathbb{N_{NF}} \ \, \mbox{if} \ \, cl(\varphi) \in \mathbb{H} \ \
\mbox{then}  \ \ \varphi \in \mathbb{{H}_{NF}}.$$
\end{theorem}

\begin{niceproof} See Subsection \ref{sect:HNCH*NC}.
\end{niceproof}

\begin{example}  \label{ex:HNC} For $ \varphi_1 $
and  $ \varphi_2 $ from   Example \ref{exam:simple},       $cl(\varphi_1) \in \mathbb{H}$ and 
 $cl(\varphi_2) \notin \mathbb{H}$;   only $\varphi_1$ is HNF.
 For $ \varphi $ and $ \varphi' $ from Example
 \ref{Ex:morecomplex},     $cl(\varphi) \in \mathbb{H}$ and   $cl(\varphi') \notin \mathbb{H}$;   
so only $ \varphi $ is HNF. 
\end{example}

Next Theorem \ref{Theorem:last} puts together previous Theorems \ref{theorem:first} and \ref{prop:sem-NNF} and provides a
concise definition of the class of HNF formulas.

\begin{theorem} \label{Theorem:last} The next statement holds:
$ \forall \varphi \in \mathbb{N_{NF}}: \  \varphi \in \mathbb{{H}_{NF}}$ \ {\em iff} \ $cl(\varphi) \in \mathbb{H}.$
\end{theorem}

\noindent {\bf Fig. 3}   exemplifies Theorem
 \ref{Theorem:last}. The classes
  $\mathbb{{H}_{NF}}$,  
$\mathbb{H}$, $\mathbb{N_{NF}}$  and   $\mathbb{C}$ are depicted 
and    each   $\varphi$ linked    to its    $cl(\varphi)$  with either  
 a red or a blue line.  
{\bf  Red  lines:}   $cl(\varphi_1)$ to 
 $cl(\varphi_3)$   are         in $\mathbb{H}$; 
     so $\varphi_{1}$ to $\varphi_{3}$ are   HNF.  
   {\bf Blue  lines:}      $cl(\phi_1)$ to 
 $cl(\phi_3)$   are  not  
  in $\mathbb{H}$; so   
  $\phi_1$ to $\phi_3$ are not HNF.

\vspace{-.3cm}

\begin{center}
\usetikzlibrary{calc,through}
\begin{tikzpicture}[scale=1.2]

\draw    [line width=.7mm] 
(-3.3,-2.1)  node[below left]  {} --
(-3.3,2.1) node[above left] {} --
(3.6,2.1) node[above right] {} --
(3.6,-2.1) node[below right] {} --  cycle;

\coordinate  [label=left:{\normalsize  \bf H}]  (N) at (.4,-.5);
\coordinate  [label=left:{\normalsize \bf C}]  (N) at (-1.,1.4);
\coordinate  [label=left:{\large \bf H{\tiny {NF}}}]  (N) at (2.,-.5);
\coordinate  [label=left:{\normalsize  \bf N{\tiny NF}}]  (N) at (-2.2,1.4);

\draw [red] [line width=.4mm] (.55,1.3) -- (1.5,1.3);
\coordinate [label=center:{\small $cl(\varphi_1)$}] (E1) at (.1,1.3);
\coordinate [label=center:{\small $\varphi_1$ \ \ \ \ }] (E1) at (2.,1.3);

\draw [blue] [line width=.4mm] (-.95,.8) -- (2.95,.8);
\coordinate  [label=center:{\ \ \ \ \small $\phi_{1}$}]  (F2) at (2.95,.8); 
\coordinate  [label=center:{\ \ \ \ \small $cl(\phi_{1})$}]  (F2) at (-1.65,.8);

\draw [red] [line width=.4mm] (.55,.2) -- (1.5,.2);
\coordinate [label=center:{\small $cl(\varphi_2)$}] (E1) at (.1,.2);
\coordinate [label=center:{\small $\varphi_2$ \ \ \ \ }] (E1) at (2.,.2);

\draw [blue] [line width=.4mm] (-2.7,-.2) -- (-1.55,-.2);
\coordinate  [label=center:{\ \ \ \ \small $cl(\phi_{2})$}]  (F2) at (-1.3,-.2);
\coordinate  [label=center:{\ \ \ \ \small $\phi_{2}$}]  (F2) at (-3.15,-.2);

\draw [red] [line width=.4mm] (.55,-1.6) -- (1.5,-1.6);
\coordinate [label=center:{\small $cl(\varphi_3)$}] (E1) at (.1,-1.6);
\coordinate [label=center:{\small $\varphi_3$ \ \ \ \ }] (E1) at (2.,-1.6);

\draw [blue] [line width=.4mm] (-.95,-1.1) -- (2.9,-1.1);
\coordinate  [label=center:{\ \ \ \ \small $\phi_{3}$}]  (F2) at (2.95,-1.1);
\coordinate  [label=center:{\ \ \ \ \small $cl(\phi_{3})$}]  (F2) at (-1.65,-1.1);

\draw    [line width=.4mm] 
(-.5,-2.)  node[below left]  {} --
(-.5,2.) node[above left] {} --
(.7,2.) node[above right] {} --
(.7,-2.) node[below right] {} --  cycle;

\draw    [line width=.4mm] 
(-2.,-2.)  node[below left]  {} --
(-2.,2.) node[above left] {} --
(.7,2.) node[above right] {} --
(.7,-2.) node[below right] {} --  cycle;

\draw    [line width=.4mm] 
(-.5,-2.)  node[below left]  {} --
(-.5,2.) node[above left] {} --
(2.5,2.) node[above right] {} --
(2.5,-2.) node[below right] {} --  cycle;
\end{tikzpicture}

\vspace{-.1cm}
{\small {\bf Fig. 3.} Exemplifying  Theorem \ref{Theorem:last}.}
\end{center}

\vspace{-.2cm}
 Theorem \ref{Theorem:last}  gives  indeed  a concise  definition of  
 $\mathbb{H_{NF}}$ but recognizing HNFs through Theorem \ref{Theorem:last} is unfeasible as
computing  
 $cl(\varphi)$ takes   exponential  time and    space. 
However   the detailed syntactical  Definition \ref{cor:syntacticalNNF}   of $\mathbb{H_{NF}}$   will allow  us (Section \ref{sec:semanticalsyntactical})  
 to   design an algorithmic strategy  to recognize HNFs in linear time.

\subsection{The Horn-NC Formulas}

  Next   define    $ \mathbb{H_{NC}} $  by simply applying  De Morgan's laws  
   to Definition \ref{theorem:visual}  of $ \mathbb{H_{NF}} $.

\begin{definition} \label{def:visual-NNC} An NC formula $\varphi$ is in $\varphi \in \mathbb{H_{NC}}$ if: (A) all its  disjunctions under the scope of no $\neg$ connective or of an even number of $\neg$ connectives 
have at most one non-negative disjunct; and (B) all its conjunctions  under the scope of an odd number of $\neg$ connectives
have at most one non-positive conjunct.
\end{definition}

\begin{example} One can check that  $ \varphi_2 $ from Example \ref{ex:NNF-NC} is HNC.  
\end{example}


\begin{definition}  \label{def:H*NC}  We define the set of formulas
 $\mathbb{\overline{H}_{NC}}$  
    exclusively from the rules below:    
\begin{itemize}
 \item     $\mathbb{{H}_{NF}} \subset  \mathbb{\overline{H}_{NC}}$.

\item    If \ $(\vee \  \varphi_{1}   \ldots \varphi_{k-1}   \,\varphi_k) \in 
\mathbb{\overline{H}_{NC}}$ 
 \  then \
  \,$\neg \{\wedge  \ \neg \varphi_1 \ldots \neg \varphi_{k-1}   \,\neg \varphi_k\} \in \mathbb{\overline{H}_{NC}}$.

\item    If \ $\{\wedge \  \varphi_{1}   \ldots \varphi_{k-1}   \,\varphi_k\} 
\in \mathbb{\overline{H}_{NC}}$    \   then \  
  \,$\neg (\vee \ \neg \varphi_1 \ldots \neg \varphi_{k-1}   \,\neg \varphi_k) \in 
  \mathbb{\overline{H}_{NC}}$.

\end{itemize}
\end{definition}

\begin{theorem} \label{Th:last}  We have that: \ $\mathbb{{H}_{NC}}=\mathbb{\overline{H}_{NC}}$.

\end{theorem}

\begin{niceproof} It follows   from   
Definition \ref{def:visual-NNC}  of $\mathbb{{H}_{NC}}$ and Definition \ref{def:H*NC} of $\mathbb{\overline{H}_{NC}}$,  
indeed both result from applying 
De Morgan's laws to $\mathbb{{H}_{NF}}$. 
\end{niceproof}

\begin{example} \label{ex:last} Applying      De Morgan's laws to the HNF 
$\{\wedge    \ \ C \  \
  \{\wedge  \ \   (\vee  \  \ A \  \ \overline{B} \,)  
 \ \  \{\wedge  \ \ \overline{A} \  \  \overline{C} \, \} \, \} \, \}$ yields 
      $\varphi_2$ in 
Example \ref{ex:NNF-NC}   (ruling out {\bf F}); so   $\varphi_2$    is $\mathrm{{HNC}}$.  
 \qed
\end{example}

\begin{definition} \label{def:HNCclass} For every $\varphi \in \mathbb{N_C}$,  
we define $cl^*(\varphi)$ as the unique clausal formula that results
 from  applying   De Morgan's laws  and   $\wedge/\vee$ distributivity   to $\varphi$ until a clausal
formula, viz. $cl^*(\varphi)$, is obtained.
\end{definition}

 Theorem  \ref{theor:charac-HNC}   syntactically and  semantically characterizes  $ \mathbb{H_{NC}} $
just as Theorems  \ref{theorem:first}, Theorem  \ref{prop:sem-NNF} and Corollary \ref{cor:firstcorprimer}  characterized  $ \mathbb{H_{NF}} $.

\begin{theorem} \label{theor:charac-HNC}
 We have that the next relationships between  $\mathbb{H_{NC}} $ and  $\mathbb{N_{NC}}$ 
 and  $\mathbb{H}$:
\begin{itemize}

\item      $\forall \varphi \in \mathbb{N_{NC}}:$ \   $\varphi \in \mathbb{H_{NC}}$ \ iff \ $cl^*(\varphi) \in \mathbb{H}$.    

\item   $\mathbb{H_{NC}} $ and  $\mathbb{H} $ are semantically  equivalent. 

\end{itemize}
\end{theorem}

\begin{niceproof} As  Definition \ref{def:HNCclass} of $cl^*(\varphi)$ is obtained  by just 
applying  De Morgan's laws to Definition \ref{def:ECNF} of $cl(\varphi)$, 
the  statements in Theorem \ref{theor:charac-HNC}   follow immediately from   
the same ones proved for $\mathbb{H_{NF}} $ in  Theorem  \ref{theorem:first},   Theorem \ref{prop:sem-NNF} and Corollary \ref{cor:firstcorprimer}, respectively. 
\end{niceproof}



\subsection{Formal Proofs} \label{sect:HNCH*NC}

Before proving Theorems  \ref{theorem:first} and \ref{prop:sem-NNF},   
Theorem  \ref{th:kdisjuncts} is required. 
So, we prove successively:
    Theorem \ref{th:kdisjuncts},
     Theorem   \ref{theorem:first} and
    Theorem \ref{prop:sem-NNF}.

\begin{theorem} \label{th:kdisjuncts}   Let  $\varphi$ be an  NNF disjunction 
$(\vee  \ \varphi_{1}   \ldots \varphi_i \ldots \varphi_k \,)$. 
   $cl(\varphi) \in \mathbb{H}$    iff  $\varphi$ has  $k-1$  negative  disjuncts and one
disjunct s.t. $cl(\varphi_i) \in \mathbb{H}$, 
 formally:  
$$cl(\,(\vee  \ \varphi_{1}   \ldots \varphi_i \ldots \varphi_k \,)\,) 
\in \mathbb{H} \mathrm{ \ \ iff \ \ } 
 (1) \  \exists i,   \mbox{\,s.t.}
  \ cl(\varphi_i) \in \mathbb{{H}} \   \mbox{\,and}   
\ \,(2) \ \forall j,  j \neq i, \varphi_j \in \mathbb{N_G}.$$    
\end{theorem}

\begin{niceproof} {\em If-then.} By refutation: let     
 $cl(\,(\vee  \ \varphi_{1}   \ldots  \varphi_i \ldots  \varphi_k)\,) \in  \mathbb{H}$ and prove that if   (1) or (2)  are  violated, then  $cl(\varphi)  \notin \mathbb{H}$. 
\begin{enumerate}
\item   [$\bullet$]   (1) $ \exists i$ s.t. $cl(\varphi_i) \notin \mathbb{{H}} $.  

 $-$ If we take  the case $k=1$, then $\varphi = \varphi_1$.

 $-$  But  $cl(\varphi_1) \notin \mathbb{H}$  implies 
  $cl(\varphi) \notin  \mathbb{H}$.

\item  [$\bullet$] (2) $\exists j,  j \neq i, \varphi_j \notin \mathbb{N_G}$. 

$-$  Suppose that, besides $\varphi_i$,   one $\varphi_j,   \,j \neq i$,   has positive literals too.

$-$  We take a simple case, concretely $k=2, \,\varphi_1=A$
and  $\varphi_2=B$.

$-$ So, $\varphi=(\vee \ \varphi_1 \ \varphi_2) = (\vee \  A  \ B)$, which implies     $cl(\varphi)  \notin \mathbb{H}$.

\end{enumerate}

$\bullet$  {\em Only-If.} For simplicity and without loss of generality, we assume  that  

\hspace{1.8cm}$(\vee \ \varphi_1 \ldots \varphi_{i} \ldots \varphi_{k-1})=\varphi^- \in  
\mathbb{N_G} 
   \mbox{ and } \varphi_k \in \mathbb{H_{NF}}$,  
and  prove:

\vspace{.2cm} 
\hspace{1.8cm}$cl(\varphi)=cl(\,(\vee \ \  \varphi_1 \ldots \varphi_i \ldots \varphi_{k-1}  \,\varphi_k) \,)
=cl(\,(\vee \ \varphi^- \ \varphi_k)\,) \in  \mathbb{H}.$ 

\vspace{.3cm}
\vspace{.0cm} $-$ To obtain $ cl(\varphi) $, one must obtain first $cl(\varphi^-)$
and $cl(\varphi_k)$, and so

 \vspace{.2cm}
\hspace{1.cm}
 $(i) \ \ cl(\varphi) = cl(\,(\vee \ \  \varphi^-  \  \varphi_k) \,)= cl(\,(\vee \ \ cl(\varphi^-)  \ \ cl(\varphi_k)\,)\,).$

\vspace{.3cm}
 $-$  By definition of $\varphi^- \in \mathbb{N_G}$, 
 
 \vspace{.2cm}
\hspace{0.8cm} $(ii) \ \  cl(\varphi^-)=\{\wedge \ D^-_1   \ldots D^-_{m-1} \,D^-_m\}$; \          
the $D^-_i$'s are negative clauses.

\vspace{.3cm}
  $-$  By definition of $\varphi_k \in \mathbb{H_{NF}}$, 

\vspace{.2cm}
\hspace{.8cm} $(iii) \ \ cl(\varphi_k)=\mathrm{H}=\{\wedge \ h_1 \  \ldots h_{n-1} \,h_n \}$; \
  the $h_i$'s are Horn clauses.

\vspace{.3cm}
 $-$ By   ($i$) to ($iii$), \ 
  
\vspace{.2cm} 
\hspace{1.cm} $cl(\varphi) = cl(\,(\vee \ \  \{\wedge \ D^-_1 \   \, \ldots \, D^-_{m-1} \,D^-_m\} \ \ \{\wedge \ h_1 \   \ldots h_{n-1} \,h_n \, \}\,)\,).$

\vspace{.3cm}
 $-$ Applying $\vee\//\wedge$ distributivity to $cl(\varphi)$ and noting $C_{i}=(\vee \ D^-_1 \ h_{i} \,)$, 

\vspace{.2cm}
\hspace{1.cm} $cl(\varphi) = 
 cl(\,\{\wedge \ \ \{\wedge \ C_1   \ldots C_i \ldots C_n\} \ \  
    (\vee \  \, \{\wedge \ D^-_2 \ldots D^-_{m-1}  \,D^-_m \, \} \  \ \mathrm{H} \,)  \,\} \ ).$

\vspace{.3cm}
 $-$ Since the  $C_i=(\vee \ D^-_1 \ h_{i} \,)$'s are  Horn clauses,  

\vspace{.2cm} 
\hspace{1.cm} $cl(\varphi) = cl (\,\{\wedge \ \ \mathrm{H}_1 \ \ 
(\vee \ \ \{\wedge \ D^-_2 \ldots D^-_{m-1} \,D^-_m \, \}   \ \ \mathrm{H}\, ) \,\} \ ).$

\vspace{.3cm}
 $-$ For $j < m$ we have,
 
\vspace{.2cm} 
\hspace{1.cm} $cl(\varphi) = cl( \ \{\wedge \ \mathrm{H}_1 \ \ldots  \mathrm{H}_{j-1}  \mathrm{H}_j   \  \ (\vee \ \ \{\wedge \ D^-_{j+1} \ldots D^-_{m-1} \,D^-_m\}  \ \  \mathrm{H} \,) \, \} \ ).$

\vspace{.3cm}
 $-$ For $j = m$, \ 
   $cl(\varphi) = \{\wedge \ \mathrm{H}_1 \  \ldots \mathrm{H}_{m-1} \,\mathrm{H}_m \     
\mathrm{H} \,\}=\mathrm{H}' \in \mathbb{H}.$

\vspace{.3cm}
 $-$  Hence   $cl(\varphi)   \in  \mathbb{H}$.
\end{niceproof}

\vspace{.1cm}

{\em 
\noindent {\bf Theorem  \ref{theorem:first}.}
We have $\forall \varphi \in  \mathbb{{H}_{NF}}:  cl(\varphi) \in \mathbb{H} $.  
}

\begin{niceproof} 
 We consider Definition \ref{cor:syntacticalNNF}  of $ \mathbb{{H}_{NF}} $. The proof 
    is done by  structural induction  on   
    the depth  $r(\varphi)$ 
   of any $\varphi$ in  $\mathbb{{H}_{NF}}$, formally defined 
  below, where $ \ell_i $ is a literal:
\[r(\varphi)= \left\{
\begin{array}{l l l}

  0   &  \varphi=\langle \odot \  \ell_1 \ \ldots \ell_{k-1} \  \ell_k \rangle \ \mbox{or} \ \varphi=\ell.\\
  
1+max\,\{r(\varphi_1), \ldots, r(\varphi_{k-1}),\,r(\varphi_k)\}    &  \varphi=\langle \odot \  \varphi_1 \ \ldots \varphi_{k-1} \  \varphi_k \rangle. \\

\end{array} \right. \]

$\bullet$ {\it Base Case:} $r(\varphi)=0.$ 

\vspace{.25cm}
\quad   \     If    $r(\varphi)=0$ clearly $\varphi \in \mathbb{H} $, and  so  $cl(\varphi) \in \mathbb{H}$.

\vspace{.3cm} 
$\bullet$ {\em Induction hypothesis:} \quad 
$     \forall \varphi,  \ r(\varphi) \leq n, \ \ \varphi \in \mathbb{{H}_{NF}} \mbox{ \ entails \ } 
cl(\varphi) \in \mathbb{H}.$

\vspace{.2cm}
$\bullet$ {\em Induction proof: \ $r(\varphi)=n+1$.}  
%
   By Definition \ref{cor:syntacticalNNF}, lines (2) and (3), we have:

\vspace{.35cm} 
\hspace{.35cm}  (2)    $\varphi=
\{\wedge  \ \varphi_1   \ldots \varphi_{i}  \ldots  \varphi_k\}
 \in \mathbb{{H}_{NF}}$, where $k \geq 1$.

  \vspace{.3cm} 
\hspace{1.2cm} $-$  By definition of $r(\varphi)$, \ $r(\varphi)=n+1$ \,entails \   $1 \leq i \leq k, \ r(\varphi_i) \leq n.$


\vspace{.25cm}
\hspace{1.2cm} $-$  By induction  hypothesis, \  $\varphi_i \in \mathbb{{H}_{NF}} 
\mbox{\, and  \,} r(\varphi_i) \leq n
  \mbox{  \ entail \ }  cl(\varphi_i) \in \mathbb{H}$.


\vspace{.25cm}
\hspace{1.2cm}  $-$ It is obvious that, \ 
%
  $  cl(\varphi) = \{\wedge  \ \  cl(\varphi_1) \  
 \ldots  \ cl(\varphi_{i})   \ldots  cl(\varphi_k) \,\}.$
    
 \vspace{.25cm}
\hspace{1.2cm} $-$  Therefore, \ 
%
$cl(\varphi) =  \{\wedge  \ \mathrm{H}_1 
\ldots \mathrm{H}_{i}  \ldots \mathrm{H}_k\}= \mathrm{H} \in \mathbb{H}$.

\vspace{.35cm}
\hspace{.35cm}    (3)
 $\varphi=(\vee  \ \varphi_1  \ldots \varphi_{i} \ldots \varphi_{k-1} \ \varphi_k) 
 \in \mathbb{{H}_{NF}}$,
where $k \geq 1$.

 \vspace{.25cm}
\hspace{1.2cm}  $-$ By  Definition \ref{cor:syntacticalNNF},  
line (3),
%
\ $0 \leq i \leq k-1, \ \varphi_i   \in \mathbb{N_G}$ and   \,$\varphi_k \in \mathbb{{H}_{NF}}.$

 \vspace{.25cm} 
\hspace{1.2cm} $-$ By  definition of $r(\varphi)$, \
%
$r(\varphi) = n+1$ \ entails   \,$r(\varphi_k) \leq n.$

 \vspace{.25cm} 
\hspace{1.2cm} $-$  By  induction  hypothesis,
%
\ $d(\varphi_k)  \,\leq n$ \,and \,$\varphi_k \,\in \mathbb{{H}_{NF}}$
 \,entail  $cl(\varphi_k) \,\in \mathbb{H}.$

\vspace{.25cm}
\hspace{1.2cm} $-$ By Theorem  \ref{th:kdisjuncts}, {\em only-if,}  
%
   $0 \leq i \leq k-1, \ \varphi_i   \in \mathbb{N_G}$ and $cl(\varphi_k) \in \mathbb{H}$ entail:

\vspace{.25cm}
\hspace{3.8cm}
 $cl(\,(\vee  \ \varphi_1  \ldots \varphi_{i} \ldots \varphi_{k-1} \ \varphi_k)\, )  \in \mathbb{H}$.
\end{niceproof}


{\em  
\noindent {\bf Theorem \ref{prop:sem-NNF}}.  $\forall \varphi \in \mathbb{N_{NF}}$:    
if  $cl(\varphi) \in \mathbb{H}$  then    $\varphi \in  \mathbb{{H}_{NF}}$.
}

\begin{niceproof} It is  done by  structural induction 
on   the depth $d(\varphi)$ of $ \varphi $   defined as follows:
\[d(\varphi)= \left\{
\begin{array}{l l l}

  0     &   \varphi \in \mathbb{C}.\\
1+max\,\{d(\varphi_1), \ldots, d(\varphi_{i}),\ldots \,,d(\varphi_k)\}   &  
\varphi=\langle \odot \  \varphi_1  \ldots \varphi_{i} \ldots  \varphi_k \rangle.\\

\end{array} \right. \]

\vspace{.2cm}
$\bullet$ {\it Base case:} $d(\varphi)=0$ and    $cl(\varphi) \in \mathbb{H}$.

\vspace{.2cm}
\ \  \  $-$  $d(\varphi)=0$ entails  $\varphi \in \mathbb{C}$.

\vspace{.2cm}
\ \  \    $-$ If $\varphi \notin \mathbb{H}$, then $cl(\varphi)  \notin  \mathbb{H}$,  
 contradicting  the  assumption.

\vspace{.2cm}
\ \  \  $-$ Hence  $\varphi \in \mathbb{H}$ and so by Definition \ref{cor:syntacticalNNF},
 $\varphi \in \mathbb{{H}_{NF}}$.


\vspace{.2cm}
$\bullet$ {\em Inductive hypothesis:} \ 
$ \forall \varphi \in \mathbb{N_{NF}}, \ d(\varphi) \leq n, \ 
cl(\varphi) \in \mathbb{H} \mbox{ \ entail \ } \varphi \in \mathbb{{H}_{NF}}.$

\vspace{.2cm}
$\bullet$ {\em Induction proof: $d(\varphi)=n+1$.} 

\vspace{.25cm}
\quad   By Definition \ref{def:NNFformulas} of $\mathbb{N_{NF}}$, cases  $(i)$ and $(ii)$  below arise.

\vspace{.3cm}
\hspace{.35cm}    $(i)$  \   $cl(\varphi)= 
  cl(\, \{\wedge \  \varphi_{1}   \ldots \varphi_{i} \ldots   \,\varphi_k\} \, )
\in \mathbb{H}$ and $k \geq 1$.

\vspace{.25cm} 
\hspace{1.2cm} $-$ Since $\varphi$ is a conjunction, $1 \leq i \leq k, \,cl(\varphi_i)  \in \mathbb{H}.$
 

 \vspace{.25cm} 
\hspace{1.2cm} $-$  By  definition of $d(\varphi)$,
%
$d(\varphi) = n+1$ \ entails   \,$1 \leq i \leq k, \,d(\varphi_i) \leq n.$

\vspace{.25cm} 
\hspace{1.2cm}  $-$ By induction,
%
\ $1 \leq i \leq k, \ \,d(\varphi_i) \leq n,  \ cl(\varphi_i) \in \mathbb{H}$
$  \mbox{\ entails\ } \varphi_i \in \mathbb{{H}_{NF}}.$

\vspace{.25cm}
\hspace{1.2cm} $-$  By Definition  \ref{cor:syntacticalNNF}, line (2),  
%
\  $1 \leq i \leq k, \ \varphi_i   \in \mathbb{{H}_{NF}}$ 
  entails $\varphi \in \mathbb{{H}_{NF}}$.

\vspace{.25cm}
 \hspace{.35cm}     $(ii)$   \ $cl(\varphi) = 
 cl(\, (\vee  \ \varphi_{1}  \ldots \varphi_i  \ldots   \varphi_{k-1} \ \varphi_k) \,) \in \mathbb{H}$ 
  and $k \geq 1$.

 \vspace{.25cm} 
\hspace{1.2cm} $-$  By   Theorem \ref{th:kdisjuncts}, {\em if-then,}  
%
\ $0 \leq i \leq k-1, \ \varphi_i  \in \mathbb{N_G}$ and   \,$cl(\varphi_k) \in \mathbb{H}.$
 
    \vspace{.25cm} 
\hspace{1.2cm} $-$  By  definition of $d(\varphi)$,
%
\ $d(\varphi) = n+1$ \ entails   \,$d(\varphi_k) \leq n.$

 \vspace{.25cm} 
\hspace{1.2cm}  $-$ By       induction  hypothesis,
%
\ $d(\varphi_k)  \,\leq n$ and $cl(\varphi_k) \in \mathbb{H}$
\,entail   $\varphi_k \,\in \mathbb{{H}_{NF}}.$

\vspace{.25cm}
\hspace{1.2cm}   $-$  By Definition \ref{cor:syntacticalNNF},  line (3),
%
  \ $0 \leq i \leq k-1, \ \varphi_i   \in \mathbb{N_G}$ and  
   \,$\varphi_k \in \mathbb{{H}_{NF}}$ 
   entail:
  
 \vspace{.25cm} 
  \hspace{3.8cm} \quad $(\vee  \ \varphi_{1}  \ldots \varphi_i  \ldots   \varphi_{k-1} \ \varphi_k) = 
  \varphi \in \mathbb{{H}_{NF}}$. 
\end{niceproof}

\section{NC Unit-Resolution   and the Tractability of   $\mathbb{{H}_{NC}}$ } \label{sec:polynomial}

We  define   {\em Non-Clausal Unit Resolution,}  noted UR$_{\mbox{nc}}$
 and  prove that,  both,  UR$_{\mbox{nc}}$ is  complete
  for  $\mathbb{{H}_{NF}}$ and that the satisfiability  of  $\mathbb{{H}_{NF}}$ is tested polynomially.
The calculus UR$_{\mbox{nc}}$
encompasses the main  rule,  called UR, and  several simplification rules. Albeit
 the latter  are straightforward, 
 UR is quite elaborate and so is presented    progressively. 

\vspace{.1cm}
 {\bf Remark.} If  $  \varphi$ is a disjunctive HNF with 
 $ k\geq 2 $ disjuncts,
then,  by Definition \ref{theorem:visual}, $  \varphi$ has at least one negative disjunct, and so,   assigning 0 to all its propositions
satisfies $  \varphi$. Hence, to avoid such simple case   we will consider that the input $  \varphi$ is a \underline{conjunctive} HNF.


\vspace{.1cm}
 At first, we introduce UR  just for   almost-clausal HNFs. 
 Assume  HNF formulas with the  almost-clausal pattern shown below and in which 
 $ {\textcolor{red} {\ell}} $ and $ {\textcolor{blue} {\overline{\ell}}} $ are any literal
 and its negated one,  and the $ \varphi $'s and the $ \phi $'s are formulas:  
$$\{{\textcolor{red} {\wedge}}  \ \varphi_1  \, \ldots  \, \varphi_{l-1} 
 \ {\textcolor{red} {\ell}} \  \varphi_{l+1}  \, \ldots  \, \varphi_{i-1} \
({\textcolor{blue} {\vee}} \ \, \phi_1 \, \ldots \, \phi_{j-1} \ 
{\textcolor{blue} {\overline{\ell}}} \ \phi_{j+1} \, \ldots \, \phi_k) \ \varphi_{i+1} \, \ldots \, \varphi_n \}$$

\noindent These formulas are almost-clausal because   
if the $\varphi$'s and   $\phi$'s were clauses and  literals, respectively, then 
they would  be  clausal.  Trivially    
such formulas are equivalent to the next ones:  
$\{{\textcolor{red} {\wedge}}  \ \varphi_1  \, \ldots  \, \varphi_{l-1} 
 \ {\textcolor{red} {\ell}} \  \varphi_{l+1}  \, \ldots  \, \varphi_{i-1} \ 
({\textcolor{blue} {\vee}} \  \phi_1 \, \ldots \, \phi_j  
\, \phi_{j+1} \, \ldots \, \phi_k) \ \varphi_{i+1} \, \ldots \, \varphi_n \}$. Thus, one can derive   the     simple   inference rule on the left, below.  By noting  $\mathcal{D}({\textcolor{blue} {\overline{\ell}}})=(\vee \ \, \phi_1 \, \ldots \, \phi_j \, \phi_{j+1} \, \ldots \, \phi_n)$,
 such rule    can be concisely written as indicated on the right:   
$$\frac{{\textcolor{red} {\ell}}     \ {\textcolor{red} {\wedge}} \
({\textcolor{blue} {\vee}} \ \ \phi_1 \ \ldots \phi_j \ 
{\textcolor{blue} {\overline{\ell}}} \ \phi_{j+1} \ \ldots \ \phi_k) }
{({\textcolor{blue} {\vee}} \ \ \phi_1 \ \ldots \phi_j  
 \ \phi_{j+1} \ \ldots \ \phi_k) } {{\mbox{\,UR}}} \qquad \quad \frac{ {\textcolor{red} {\ell}} \ \, {\textcolor{red}\wedge} \ \, 
({\textcolor{blue} {\vee}} \ \, {\textcolor{blue} {\overline{\ell}}} \ \, \mathcal{D}({\textcolor{blue} {\overline{\ell}}}) \,) }
{  \mathcal{D}({\textcolor{blue} {\overline{\ell}}})  }{{\mbox{\,UR}}}$$


\noindent   We now extend our analysis from formulas with the previous pattern 
 ${\textcolor{red} {\ell}} \, {\textcolor{red}\wedge} \,  
({\textcolor{blue} {\vee}} \ \, {\textcolor{blue} {\overline{\ell}}} \ \, \mathcal{D}({\textcolor{blue} {\overline{\ell}}}) \,)$ 
to those with  pattern  
 ${\textcolor{red} {\ell}}  \, {\textcolor{red} {\wedge}} \, ({\textcolor{blue} {\vee}} \   
\mathcal{C}({\textcolor{blue} {\overline{\ell}}})  \  \mathcal{D}({\textcolor{blue} {\overline{\ell}}})\, )$ 
where $\mathcal{C}({\textcolor{blue} {\overline{\ell}}})$
is \underline{the maximal sub-formula linked to ${\textcolor{red} {\ell}}$} 
  that becomes false when $ {\textcolor{blue} {\overline{\ell}}} $ is false, i.e.
$\mathcal{C}({\textcolor{blue} {\overline{\ell}}})$ is  the maximal  sub-formula
    equivalent to a conjunction  
  $ {\textcolor{blue} {\overline{\ell}}} \wedge \psi$.
%
   For instance, if $\varphi  $ has the sub-formula 
   $(\vee \ \varphi_1 \  \{\wedge \ \phi_ 1 \ \{\wedge \ {\textcolor{blue} {\overline{\ell}}} \ 
 (\vee \ \phi_2 \ \overline{A}) \ \phi_3\} \}\ \varphi_2\,)$, we take  
 $\mathcal{C}({\textcolor{blue} {\overline{\ell}}})=\{\wedge \ \phi_ 1 \ \{\wedge \ {\textcolor{blue} {\overline{\ell}}} \ 
 (\vee \ \phi_2 \ \overline{A}) \ \phi_3\} \}$ 
 because: (a)   $\mathcal{C}({\textcolor{blue} {\overline{\ell}}})$ verifies ${\textcolor{blue} {\overline{\ell}}} \wedge \psi= {\textcolor{blue} {\overline{\ell}}} 
 \wedge \{\wedge \ \phi_1 \   \, (\vee \ \phi_2 \ \overline{A}) \ \, \phi_3  \, \} $; 
 and (b) no bigger sub-formula verifies (a). 
 It is easily checked that, 
the  augmented  UR  for the pattern  
$ {\textcolor{red} \ell} \ \, {\textcolor{red}\wedge} \ \,
({\textcolor{blue} {\vee}} \ \, \mathcal{C}({\textcolor{blue} {\overline{\ell}}}) \ \, \mathcal{D}({\textcolor{blue}{\overline{\ell}}}) \,) $ is $ {\textcolor{red} \ell} \ \, {\textcolor{red}\wedge} \ \,
({\textcolor{blue} {\vee}} \ \, \mathcal{C}({\textcolor{blue} {\overline{\ell}}}) \ \, \mathcal{D}({\textcolor{blue}{\overline{\ell}}}) \,) \vdash \mathcal{D}({\textcolor{blue}{\overline{\ell}}})$.

\begin{example} \label{ex:simple-conjuction}  Let us consider the formula 
below where the $ \phi_i $'s
are HNF formulas:
$$ \varphi=  \{{\textcolor{red} {\wedge}} \  \ {\textcolor{red} A} \ \    
(\vee   \ \  \overline{C} \ \  \{\wedge \ \ {\textcolor{blue} {\overline{A}}} \ \ (\vee \ \ E  \ \ \overline{D} ) \, \} \ \ \phi_1\,) \ \ 
(\vee \ \ D \ \   \{\wedge \ \ \phi_2  \ \ \overline{A} \,\} \, ) \ \ \phi_3 \,\}.$$
If we take   the left-most
 ${\textcolor{blue} {\overline{A}}}$, then $\varphi$ has a sub-formula with pattern  
   ${\textcolor{red} {A}} \ {\textcolor{red} {\wedge}} \ ({\textcolor{blue} {\vee}} \   
\mathcal{C}({\textcolor{blue} {\overline{A}}}) \ \mathcal{D}({\textcolor{blue} {\overline{A}}})\, )$, in which  
$\mathcal{C}({\textcolor{blue} {\overline{A}}})=\{\wedge \  {\textcolor{blue} {\overline{A}}} \ \ (\vee \ E \ \overline{D} ) \, \}$ and $\mathcal{D}({\textcolor{blue} {\overline{A}}})= 
(\vee \  \overline{C} \ \phi_1).$
 By applying UR
to  $ \varphi $, we obtain:

\vspace{.3cm}
\hspace{2.5cm}$\varphi'= \{{\textcolor{red} {\wedge}} \  \ {\textcolor{red} A} \ \    
(\vee   \ \  \overline{C}  \ \phi_1 ) \ \ 
(\vee \ \ D \ \   \{\wedge \ \ \phi_2  \ \ \overline{A} \,\} \, ) \ \ \phi_3 \,\}.  $ \qed
\end{example}


\noindent  The  pattern of the   almost-clausal HNFs
 can be rewritten $\{{\textcolor{red} {\wedge}}  \ \Pi
 \ {\textcolor{red} {\ell}} \  
({\textcolor{blue} {\vee}} \   \mathcal{C}({\textcolor{blue} {\overline{\ell}}}) \  
 \mathcal{D}({\textcolor{blue} {\overline{\ell}}}) \,) \ \Pi' \}$, where $ \Pi $ and $ \Pi' $
  represent concatenations of
 formulas, i.e. $ \Pi=\varphi_1 \ldots \varphi_{i-1} $ and 
 $ \Pi'=\varphi_{i+1} \ldots \varphi_k $. 
 Then it is not hard to check that the   general nested HNFs to which  the rule UR  can indeed  be applied,
  must have  the next syntactical 
 pattern\footnote{The notation $ \langle \odot \ \varphi_1 \ldots \varphi_k \rangle $ was introduced 
 in Definition \ref{def:NNFformulas}, bottom.}:
 $$\{\wedge \ \Pi_0 \   {\textcolor{red} \ell}  
 \ \langle \odot_1 \ \, \Pi_1 \ \   \ldots  \ \   \langle \odot_k \ \, \Pi_{k} \ \,  
 ({\textcolor{blue} {\vee}} \   \mathcal{C}({\textcolor{blue} {\overline{\ell}}}) \  
 \mathcal{D}({\textcolor{blue} {\overline{\ell}}}) \,) \ \,  \Pi'_k \, \rangle \ \ 
    \ldots  \ \ \Pi'_1 \,\rangle \ \Pi_0' \, \}$$

\noindent where   the $\Pi_j$'s and  $\Pi_j'$'s are concatenations of HNF formulas, namely 
$\Pi_j=\varphi_{j_1} \ldots \varphi_{j_{i-1}}$  and 
$\Pi'_j=\varphi_{j_{i+1}} \ldots \varphi_{j_{n_j}},$ 
 and the rule   UR can be expressed
 as follows:
$$\frac{ {\textcolor{red} \ell}  
\ {\textcolor{red} {\wedge}} \  \langle\odot_1 \ \Pi_1    \ldots  \langle \odot_k  
 {\textcolor{red}  \ \ \Pi_{k} \ \  
 ({\textcolor{blue} {\vee}} \ \  \mathcal{C}({\textcolor{blue} {\overline{\ell}}}) \ \ \mathcal{D}({\textcolor{blue} {\overline{\ell}}}) \,)  
 \ \ \Pi'_k\,\rangle } \ldots  \, \Pi'_1\rangle }
 {  
 \langle\odot_1 \ \Pi_1   \ldots  \langle\odot_k  
 {\textcolor{blue}  \ \ \Pi_{k} \ \  
  \mathcal{D}({\textcolor{blue} {\overline{\ell}}})  \ \ \Pi'_k\,\rangle }
  \ldots  \, \Pi'_1 \rangle } 
 {\ \mbox{UR}}$$
 
\noindent   \underline{UR means}: if the input 
 has a literal   $ {\textcolor{red} \ell} $ conjunctively linked 
 to a sub-formula     with the pattern
of the right conjunct in the   numerator, then the latter  can be  replaced with the formula in the
  denominator. In practice, applying UR  is equivalent to   remove 
  $\mathcal{C}({\textcolor{blue} {\overline{\ell}}})$.
   If we   denote $\Pi$ the right conjunct of the numerator  and  by
$\Pi \cdot ({\textcolor{blue} {\vee}} \  \mathcal{C}({\textcolor{blue} {\overline{\ell}}}) \  \mathcal{D}({\textcolor{blue} {\overline{\ell}}})  \,)$  that $({\textcolor{blue} {\vee}} \ \mathcal{C}({\textcolor{blue} {\overline{\ell}}}) \  \mathcal{D}({\textcolor{blue} {\overline{\ell}}}) \,)$ is a sub-formula of $\Pi$, then    UR
 can be expressed:
$$\tcboxmath{\frac{{\textcolor{red} \ell} \ {\textcolor{red} {\wedge}} \ 
\Pi \cdot ({\textcolor{blue} {\vee}} \ 
 \ \mathcal{C}({\textcolor{blue} {\overline{\ell}}}) \ \ \mathcal{D}({\textcolor{blue} {\overline{\ell}}})  \,)  }
{
\Pi \cdot  \mathcal{D}({\textcolor{blue} {\overline{\ell}}})    }{\mbox{\,UR}}}$$

\begin{example} \label{ex:formulacompl} Let us consider that the following  formula is HNF: 
$$\varphi=\{\wedge \ \ (\vee \ \ C \ \phi_1) \ \ (\vee \ \ \overline{A} \ \ \{\wedge \  \   
(\vee   \ \   \overline{A} \  \ \overline{C} \,) 
\quad (\vee \ \ \phi_2 \ \   \{\wedge \ \ \overline{B}  \ \ {\textcolor{blue} {\overline{A}}} \,\} \, ) \ \ C \, \} \, ) \ {\textcolor{red} A} \,\}.$$

\noindent Its associated DAG (tree, in this is case) is depicted in {\bf Fig. 5}, left.

\vspace{-.3cm}
\begin{center}
\begin{adjustbox}{valign=t}
\begin{tikzpicture}[]
\Tree
[.$\wedge$   [.$\vee$ [.C\\ ][.$\phi_1$\\ ] ] 
             [.$\vee$  [.{\color{black}$\overline{A}$}\\ ] 
                       [.$\wedge$  [.$\vee$ [.$\overline{A}$\\ ] [.$\overline{C}$\\ ]  ] 
                                   [.$\vee$ [.$\phi_2$\\ ] 
                                              [.$\wedge$ [.$\overline{B}$\\ ] [.{\color{blue}$\overline{A}$}\\ ] ]
                                              ]
                                   [.C\\ ] 
                       ]
              ]
             [.{\color{red} A}\\ ]
        ]
\end{tikzpicture}
\end{adjustbox} 
\begin{adjustbox}{valign=t} \hspace{.5cm}
\begin{tikzpicture}[]
\Tree
[.$\wedge$   [.$\vee$ [.C\\ ][.$\phi_1$\\ ] ] 
             [.$\vee$  [.{\color{blue}$\overline{A}$}\\ ] 
                       [.$\wedge$  [.$\vee$ [.$\overline{A}$\\ ] [.$\overline{C}$\\ ]  ] 
                                   [.$\vee$ [.$\phi_2$\\ ] 
                                              ]
                                   [.C\\ ] 
                       ]
              ]
             [.{\color{red} A}\\ ]
        ]
\end{tikzpicture}
\end{adjustbox}

\vspace{-.3cm}
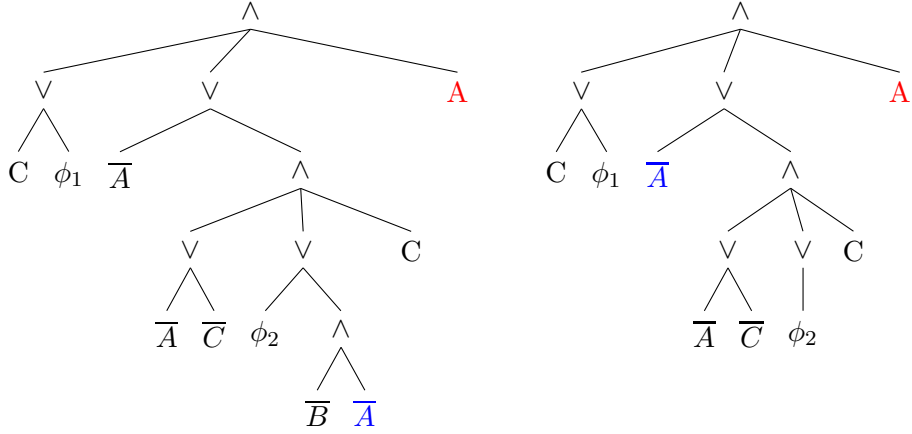
{\small {\bf Fig.  5.} {\em Formulas $ \varphi $ (left) and  $ \varphi' $ (right)}}
\end{center}


\vspace{-.2cm}
\noindent For  ${\textcolor{red} A}$ and the right-most 
${\textcolor{blue} {\overline{A}}}$,  the formulas in UR are the following:

\vspace{.2cm}
$\bullet  $ $\Pi= (\vee \ \ \overline{A} \ \ \{\wedge \  \   
(\vee   \ \   \overline{A} \  \ \overline{C} \,) 
\quad (\vee \ \ \phi_2 \ \   \{\wedge \ \ \overline{B}  \ \ {\textcolor{blue} {\overline{A}}} \,\} \, ) \ \ C \, \} \, )$
 
 \vspace{.2cm}
$\bullet  $ $ (\vee  \ \mathcal{C}({\color{blue}\overline{A}}) \ \mathcal{D}({\color{blue}\overline{A}}))
=(\vee \ \ \phi_2 \ \   \{\wedge \ \ \overline{B}  \ \ {\color{blue}\overline{A}} \,\} \, )$

 \vspace{.2cm}
 $\bullet  $ $  \mathcal{C}({\color{blue}\overline{A}})=\{\wedge \ \ \overline{B}  \ \ {\color{blue}\overline{A}} \,\}$ 
 \ \ and \ \ $  \mathcal{D}({\color{blue}\overline{A}})=\phi_2$.

\vspace{.2cm}
\noindent Then, applying   UR to $ \varphi $
leads $ \varphi' $ below and whose   tree is   the right  one in  {\bf Fig. 5}: 

\vspace{.3cm} 
\hspace{1.3cm}$\varphi'=\{\wedge \ \ (\vee \ \ C \ \phi_1) \ \ (\vee \ \ \overline{A} \ \ \{\wedge \  \   
(\vee   \ \   \overline{A} \  \ \overline{C} \,) 
\quad (\vee \ \ \phi_2  \, ) \ \ C \, \} \, ) \ {\textcolor{red} A} \,\}$ \qed
\end{example}

\vspace{.1cm}
\noindent  $\bullet$ {\bf Simplification Rules.}    
The simplification
 rules given below must  accompany  UR.  The first two rules  
  simplify  formulas by (upwards) propagating 
   $(\vee)$   from sub-formulas to formulas;
  $ \varphi $ is the input HNF, and as before,  $ \varphi \cdot \phi $ means that  $ \phi $ is a sub-formula of $ \varphi $:
  
$$\frac{  \varphi \cdot ({\textcolor{black} {\vee}} \  \ \phi_1 \ldots   \phi_{i-1} 
\, (\vee) \, \phi_{i+1} \ldots \phi_k  \,)  }
{ \varphi \cdot ({\textcolor{black} {\vee}} \  \ \phi_1 \ldots   \phi_{i-1} 
\,  \phi_{i+1} \ldots \phi_k  \,)  }{{\bf F}_\vee} \qquad \frac{
\varphi \cdot \{{\textcolor{black} {\wedge}} \  \ \varphi_1 \ldots   \varphi_{i-1} 
\, (\vee) \, \varphi_{i+1} \ldots \varphi_k  \,\}  }
{ \varphi \cdot  (\, {\textcolor{black} {\vee}} \,)  }{{\bf F}_\wedge}$$

\noindent The next two rules  remove redundant connectives; the first one 
removes a connective $ \odot $ if it is applied to a single formula,
i.e. $ \langle \odot  \  \phi_1 \,\rangle $,
and the second one removes a connective  that is inside
another equal connective:

$$\frac{ \varphi \cdot \langle{\textcolor{black} {\odot_1}} \  \ \varphi_1 \ldots   \varphi_{i-1} 
\, \langle \odot_2 \  \phi_1 \,\rangle \, \varphi_{i+1} \ldots \varphi_k  \,\rangle  }
{\varphi \cdot \langle{\textcolor{black} {\odot_1}} \  \ \varphi_1 \ldots   \varphi_{i-1} 
 \,  \phi_1 \,  \varphi_{i+1} \ldots \varphi_k  \,\rangle }{\ \odot \phi}$$
 
$$\frac{ \varphi \cdot \langle{\textcolor{black} {\odot_1}} \  \ \varphi_1 \ldots   \varphi_{i-1} 
\, \langle\odot_2 \  \phi_1 \ldots \phi_n \,\rangle \, \varphi_{i+1} \ldots \varphi_k  \,\rangle, \odot_1=\odot_2  }
{\varphi \cdot \langle{\textcolor{black} {\odot_1}} \  \ \varphi_1 \ldots   \varphi_{i-1} 
 \,  \phi_1 \ldots \phi_n \,  \varphi_{i+1} \ldots \varphi_k  \,\rangle }{\ \odot \odot}$$

\begin{example} \label{ex:cont} We continue with  Example  \ref{ex:formulacompl}.
 Picking   the left-most  ${\color{blue}\overline{A}}$ 
(colored blue  in {\bf Fig. 5}, right),   we have  $\mathcal{C}({\color{blue}\overline{A}})={\color{blue}\overline{A}}$,  \ 
 $\mathcal{D}({\color{blue}\overline{A}})=(\vee \ \{\wedge \  \   
(\vee   \    \overline{A} \  \ \overline{C} \,) 
\ \  (\vee \  \phi_2  \, ) \ \ C \, \} \, )$, and  
$\Pi  =
(\vee \ \mathcal{C}({\color{blue}\overline{A}}) \ \mathcal{D}({\color{blue}\overline{A}})) $.  
 By applying   UR to $ \varphi' $,   the obtained formula  is given  in    {\bf Fig. 6}, left.
 After two applications of  $\odot \phi$ and one of $\odot \odot$, one gets the formula associated with the  right tree in  {\bf Fig. 6.}
Finally,   two  applications of  UR  to the two  pairs {\textcolor{red} C} and {\textcolor{blue}{$\overline{C}$}}, and  {\textcolor{red} A} and {\textcolor{blue}{$\overline{A}$}},  and then one application of ${\bf F}_\wedge$
leads the calculus  UR$_{\mbox{nc}}$ to  derive $(\vee)$.

\vspace{-.2cm} 
\begin{center}
\begin{adjustbox}{valign=t}
\begin{tikzpicture}[]
\Tree
[.$\wedge$   [.$\vee$ [.C\\ ][.$\phi_1$\\ ] ] 
             [.$\vee$   
                       [.$\wedge$  [.$\vee$ [.$\overline{A}$\\ ] [.$\overline{C}$\\ ]  ] 
                                   [.$\vee$ [.$\phi_2$\\ ] 
                                              ]
                                   [.C\\ ] 
                       ]
              ]
             [.{\color{red} A}\\ ]
        ]
\end{tikzpicture}
\end{adjustbox}
\begin{adjustbox}{valign=t} \hspace{2.cm}
\begin{tikzpicture}[]
\Tree
[.$\wedge$   [.$\vee$ [.C\\ ][.$\phi_1$\\ ] ]   
                        [.$\vee$ [.{\textcolor{blue}{$\overline{A}$}}\\ ] [.{\textcolor{blue}{$\overline{C}$}}\\ ]  ] 
                                    [.$\phi_2$\\ ] 
                                   [.{\textcolor{red} C}\\ ] 
             [.{\textcolor{red} A}\\ ]
        ]
\end{tikzpicture}
\end{adjustbox}

\vspace*{-.6cm}
{\small {\bf Fig. 6.} {\em Example \ref{ex:cont}}} 
\end{center}
\end{example}

\vspace{-.4cm}
\begin{definition} We  define    UR$_{\mbox{nc}}$ as the  calculus
 formed by  the UR rule  and the above described  simplification rules, i.e.  
     UR$_{\mbox{nc}}=\{\mathrm{UR},{\bf F}_\vee, {\bf F}_\wedge, \odot \phi, \odot \odot \}$.
\end{definition}

\begin{lemma} An HNF $\varphi$ is unsatisfiable iff     UR$_{\mbox{nc}}$ 
applied  to $\varphi$ derives $(\vee)$.
\end{lemma}

 The proof is rather straightforward and   contains some tedious details, and so it is
omitted by space reasons.

\begin{lemma} \label{lem:polytime}  HNF-SAT is polynomial.
\end{lemma}

\begin{niceproof} The number of   UR and simplification rules
 performed 
is at most the     number of literals and connectives
in $ \varphi $, respectively,. On the other hand, 
it is not difficult to find 
data structures to execute polynomially each rule in UR$_{\mbox{nc}}$.
Hence, the lemma holds. 
\end{niceproof}

\begin{proposition} HNF-SAT is P-complete.
\end{proposition}

\begin{niceproof} It follows straightforwardly from  HNF-SAT is polynomial, by Lemma
\ref{lem:polytime}, and HNF-SAT includes   Horn-SAT, which is P-complete \cite{DantsinEGV01, BryEEFGLLPW07}.
\end{niceproof}

\noindent {\bf Final Remark.} 
%
%
%
%
We   believe that
 it will be not    hard to find data structures to devise  an 
 HNF-SAT algorithm based on 
   UR$_{\mbox{nc}}$ with  linear complexity.
 Since the HNF-SAT algorithm is
 NC Unit-Propagation,  on its basis,   
 effective  NC DPLL solvers  can be developed.

\subsection{Beyond NC Unit-Resolution} \label{subsec:generalNNFs}


  {\bf (1) NC Hyper-Unit-Resolution.} 
The given definition of the NC Unit-Resolution rule  can be  extended  
to obtain NC Hyper-Unit-Resolution (HUR).
 Assume that the $ \varphi $ contains a conjunction 
 of $ \textcolor{red} \ell  $ with two 
 sub-formulas $ ({\textcolor{blue} {\vee}} \ 
 \ \mathcal{C}({\textcolor{blue} {\overline{\ell}^1}}) \ \ 
 \mathcal{D}({\textcolor{blue} {\overline{\ell}^1}})  \,) $ and
 $ ({\textcolor{blue} {\vee}} \ 
 \ \mathcal{C}({\textcolor{blue} {\overline{\ell}^2}}) \ \ 
 \mathcal{D}({\textcolor{blue} {\overline{\ell}^2}})  \,) $,
 where $\overline{\ell}^i $ denotes a specific occurrence
 of $ \overline{\ell} $. The simultaneous application of NC unit-resolution with two sub-formulas
 is formally expressed as follows:
$$\frac{   \textcolor{red} \ell  
\ {\textcolor{red} {\wedge}} \ 
 \Pi^1 \succ \, ({\textcolor{blue} {\vee}} \ 
 \ \mathcal{C}({\textcolor{blue} {\overline{\ell}^1}}) \ \ 
 \mathcal{D}({\textcolor{blue} {\overline{\ell}^1}})  \,)    \ 
 {\textcolor{red} {\wedge}} \   
  \Pi^2 \succ \, ({\textcolor{blue} {\vee}} \ 
 \ \mathcal{C}({\textcolor{blue} {\overline{\ell}^2}}) \ \ 
 \mathcal{D}({\textcolor{blue} {\overline{\ell}^2}})  \,) }
{   \Pi^1 \succ  \mathcal{D}({\textcolor{blue}{\overline{\ell}^1}})  
 \wedge \,
  \Pi^2 \succ  \mathcal{D}({\textcolor{blue} {\overline{\ell}^2}})    }{\mbox{\em }}$$

\noindent If   the sub-formula
 $  \Pi^i \succ ({\textcolor{black} {\vee}} \ 
  \mathcal{C}({\textcolor{black} {\overline{\ell}^i}}) 
 \  \mathcal{D}({\textcolor{black} {\overline{\ell}^i}}) ) $ 
 is denoted $ \langle \Pi, 
 \mathcal{CD}({\textcolor{black} {\overline{\ell}}})  \rangle^i$,
 then 
 HUR  for $ k $ sub-formulas is formally expressed  as follows:
 
$$\tcboxmath{\frac{ 
  {\textcolor{red} {\ell }}   \ {\textcolor{red} {\wedge}} \ 
\langle \Pi, \mathcal{CD}({\textcolor{blue} {\overline{\ell}}}) \rangle^1
{\textcolor{red} {\wedge}} \ldots {\textcolor{red} {\wedge}} \ 
\langle \Pi, \mathcal{CD}({\textcolor{blue} {\overline{\ell}}})  \rangle^i \
 {\textcolor{red} {\wedge}} \,\ldots  {\textcolor{red} {\wedge}}
\ \langle \Pi , \mathcal{CD}({\textcolor{blue} {\overline{\ell}}}) \rangle^k  }
{ \langle \Pi, \mathcal{D}({\textcolor{blue} {\overline{\ell}}})  \rangle^1 \ 
{\textcolor{red} {\wedge}} \ \ldots \ {\textcolor{red} {\wedge}} \
\langle \Pi, \mathcal{D}({\textcolor{blue} {\overline{\ell}}}) 
 \rangle^i \ {\textcolor{red} {\wedge}} \
\ldots \ {\textcolor{red} {\wedge}} \
\langle \Pi, \mathcal{D}({\textcolor{blue} 
{\overline{\ell}}}) \rangle^k  } 
{\mbox{ \ \underline{HUR}}}}$$

\noindent Since
the  $ {\textcolor{black} {\overline{\ell}^i }}$'s   
    are pairwise different, so are the   sub-formulas 
 $ \mathcal{CD}^i $ and $ \mathcal{D}^i $
in HUR. However, the formulas $ \Pi^i $
are not necessarily different as it is shown next:

  \begin{example} \label{ex:HUR}   Let us reconsider also the formula in previous Example 
  \ref{ex:formulacompl}:
$$\varphi=\{\wedge \ \ (\vee \ \ C \ \phi_1) \ \ (\vee \ \ 
{\textcolor{blue} {\overline{A}^1}} \ \ \{\wedge \  \   
(\vee   \ \   {\textcolor{blue} {\overline{A}^2}} \  \ \overline{C} \,) 
\quad (\vee \ \ \phi_2 \ \   \{\wedge \ \ \overline{B}  \ \ {\textcolor{blue} {\overline{A}^3}} \,\} \, ) \ \ C \, \} \, ) \ {\textcolor{red} A} \,\}.$$

\noindent One can apply HUR with $ {\textcolor{red} A}  $ and the three literals ${\textcolor{blue} {\overline{A}^i}}$. The formula $ \Pi $ in the numerator
of HUR is   the same for the three literals, so it is noted $ \Pi^{1,2,3}$ below, but the
formulas $ (\vee \ \ \mathcal{C}(\overline{A}^i) \ \ \mathcal{D}(\overline{A}^i))$
 are different and are as follows:

\vspace{.2cm}
-- $ \Pi^{1,2,3}= (\vee \ \ 
{\textcolor{blue} {\overline{A}^1}} \ \ \{\wedge \  \   
(\vee   \ \   {\textcolor{blue} {\overline{A}^2}} \  \ \overline{C} \,) 
\quad (\vee \ \ \phi_2 \ \   \{\wedge \ \ \overline{B}  
\ \ {\textcolor{blue} {\overline{A}^3}} \,\} \, ) \ \ C \, \} \, )$

\vspace{.15cm}
-- $ (\vee \ \ \mathcal{C}({\textcolor{blue} {\overline{A}^1}}) 
\ \ \mathcal{D}({\textcolor{blue} {\overline{A}^1}}))=\Pi^{1,2,3}$

\vspace{.15cm}
-- $ (\vee \ \ \mathcal{C}({\textcolor{blue} {\overline{A}^2}}) 
\ \ \mathcal{D}({\textcolor{blue} {\overline{A}^2}}))= 
(\vee   \ \   {\textcolor{blue} {\overline{A}^2}} \  \ \neg {R} \,) $

\vspace{.15cm}
-- $ (\vee \ \ \mathcal{C}({\textcolor{blue} {\overline{A}^3}}) 
\ \ \mathcal{D}({\textcolor{blue} {\overline{A}^3}}))= 
(\vee \ \ \phi_2 \ \   \{\wedge \ \ \neg {Q}  \ \ {\textcolor{blue} {\overline{A}^3}} \,\} \, )$

\vspace{.25cm}
\noindent Applying NC Hyper Unit-Resolution yields:
$   \{\wedge \ \ (\vee \  R \ \phi_1) \ \ 
(\vee \ \   \{\wedge \  \   
(\vee   \      \neg {R}) 
\ \ (\vee \  \phi_2) \ \ R\}) \ {\textcolor{red} P} \,\} $

\vspace{.1cm}
\noindent After simplifying:
$    \{\wedge \ \ (\vee \ \ C\ \phi_1) \ \ 
       \overline{C} \ \  \phi_2    \ \ C   \,\ {\textcolor{red} A} \,\} $. 
Clearly,   a simple NC unit-resolution deduces $   (\vee)  $.
Altogether,   the rule   HUR accelerates considerably
the proof of  unsatisfiability. \qed
\end{example}

{\bf Remark.} An NC hyper unit-resolution rule even more general than 
 ${\mbox{HUR}}$ 
 can be devised to handle simultaneously $ k\geq 1 $  
     unit-clauses, 
 $   \ell_k  $, 
      so that for each  
$   \ell_k  $,  
  $i \geq  1  $   sub-formulas $  \Pi^{k,i} \succ ({\textcolor{black} {\vee}} \ 
  \mathcal{C}({\textcolor{black} {\neg \ell^{k,i}}}) 
 \  \mathcal{D}({\textcolor{black} {\neg \ell^{k,i}}}) ) $  can be covered. 
 That is, one can compute simultaneously $ k \geq 1 $ 
 applications of    HUR   (but for space reasons, we omit its formal
 definition).

\vspace{.2cm}
\noindent {\bf (2)   NC Local-Unit-Resolution.}  
The  UR rule can also be  applied to \underline{sub-formulas} 
in order to accelerate the  searching of proofs, i.e. to allow for shorter proofs.  Thus,
if a  sub-formula $ \phi $ 
of  $ \varphi $ has  the pattern of UR, UR can be applied to $ \phi $. Namely,
applying   UR to any sub-formula of  $\varphi $  with pattern   
 $\phi={\textcolor{red} \ell} \ {\textcolor{red} {\wedge}} \ 
\Pi \cdot ({\textcolor{blue} {\vee}} \   \mathcal{C}({\textcolor{blue} {\overline{\ell}}}) \  \mathcal{D}({\textcolor{blue} {\overline{\ell}}})  \,)$    can be authorized
and    $ \phi $   replaced with ${\textcolor{red} \ell} \ {\textcolor{red} {\wedge}} \ 
\Pi \cdot     \mathcal{D}({\textcolor{blue} {\overline{\ell}}}) $. 
So,   
   Local NC Unit-Resolution, LUR, is formalized as follows:
$$\tcboxmath{\frac{\varphi \cdot (\,{\textcolor{red} \ell} \ {\textcolor{red} {\wedge}} \ 
\Pi \cdot ({\textcolor{blue} {\vee}} \  \ \mathcal{C}({\textcolor{blue} {\overline{\ell}}}) \ \ \mathcal{D}({\textcolor{blue} {\overline{\ell}}})  \,) \,) }
{\varphi \cdot (\,{\textcolor{red} \ell} \ {\textcolor{red} {\wedge}} \ 
\Pi \cdot    \mathcal{D}({\textcolor{blue} {\overline{\ell}}})  \,)  }{\mbox{\,LUR}}}$$

\noindent \underline{LUR means}: if the input $ \varphi $ has a sub-formula
with pattern $ {\textcolor{red} \ell} \ {\textcolor{red} {\wedge}} \ 
\Pi \cdot ({\textcolor{blue} {\vee}} \  \ \mathcal{C}({\textcolor{blue} {\overline{\ell}}}) \ \ \mathcal{D}({\textcolor{blue} {\overline{\ell}}})  \,) $, the
latter can be replaced with $ {\textcolor{red} \ell} \ {\textcolor{red} {\wedge}} \ 
\Pi \cdot    \mathcal{D}({\textcolor{blue} {\overline{\ell}}}) $,
or analogously, the formula $\mathcal{C}({\textcolor{blue} {\overline{\ell}}})$ can be eliminated.

\begin{example} Consider again  Example \ref{ex:formulacompl}.
 Its sub-formula
$\{\wedge \ \ (\vee \ \ \overline{A} \ \ {\textcolor{blue} {\overline{C}}} \,) 
\ \ (\vee \ \ \phi_2 \ \   \{\wedge \ \ \overline{B}  
\ \ {\textcolor{black} {\overline{A}}} \,\}\\ \, ) \ \ {\textcolor{red} C} \, \}$ has the  pattern ${\textcolor{red} \ell} \ {\textcolor{red} {\wedge}} \ 
\Pi \cdot ({\textcolor{blue} {\vee}} \   \mathcal{C}({\textcolor{blue} {\overline{\ell}}}) \  \mathcal{D}({\textcolor{blue} {\overline{\ell}}})  \,)$, and
so  LUR can be applied to $ \varphi $ as follows:
$ \varphi \cdot (\,{\textcolor{red} C} \wedge \Pi \cdot (\vee   \ \   \overline{A} \  \ {\textcolor{blue} {\overline{C}}} \,)\, ) \vdash \varphi \cdot (\,{\textcolor{red} C} \wedge \Pi \cdot   \overline{A}  \, ) $. After applying LUR, one obtains:
$$\{\wedge \ \ (\vee \ \ C \ \phi_1) \ \ (\vee \ \ {\textcolor{black} {\overline{A}}} \ \ \{\wedge \  \   
   {\textcolor{black} {\overline{A}}}  
\quad (\vee \ \ \phi_2 \ \   \{\wedge \ \ \overline{B}  \ \ {\textcolor{black} {\overline{A}}} \,\} \, ) \ \ C \, \} \, ) \ {\textcolor{black} A} \,\}.$$
 Now applying HUR with ${\textcolor{red} A}$ and the first two ${\textcolor{blue} {\overline{A}}}$'s,
one determines that $ \varphi $ is unsatisfiable.
  \qed
\end{example}

\begin{proposition}  If  
  applying LUR to any NNF formula $ \varphi $ results in $ \varphi' $ then $ \varphi $  and  $ \varphi' $ are logically equivalent.
\end{proposition}

The proof  follows straightforwardly from the soundness of UR.

\vspace{.1cm}
 Observe that  HUR and  LUR   habilitate     shorter proofs. For instance, for Example  \ref{ex:formulacompl}, only one LUR and one HUR are enough
 to derive $ (\vee) $.
Thus, 
  their
 suitable management  can make increase the overall deductive efficiency.
 
 \subsection{Application to Logic Programming} 
 \label{sub:applic-log-prorm}
 
  The introduction of NC formulas in 
  the field of logic programming, concretely 
 within answer set programming, was done in
\cite{LifschitzTT99}. Programs containing NCs 
in their bodies and heads are called nested programs
and have been employed
in many issues  e.g. normal forms \cite{BriaFL09}, extension to other
logics \cite{NievesL12}, allowing for highly-expressive predicates (aggregates) \cite{CabalarFSS19}, etc. However, nested definite  programs, the equivalent to classical definite programs,
have not been characterized. We show  that the HNF formulas
are indeed the base of the nested definite  programs.

\begin{definition} \label{def:NCrules}  An HNF  rule is an expression  $ \Pi^+ \rightarrow \Phi $ wherein $ \Pi^+ $ is an 
NNF  having only positive literals
and $\Phi$ is an arbitrary HNF. An HNF  program is a set of HNF rules.
\end{definition}

\begin{example} \label{ex:NCrule} The nest rule is an HNF rule 
because its body   is  a  positive NNF and 
 its head  is an HNF:
 
 \vspace{.2cm}
\hspace{1.cm}$\{\wedge \ A \ D \} \ (\vee \ E  \ \{\wedge \ D \ C\}) \  \longrightarrow \ \{\wedge \  \   
(\vee   \ \   \overline{A} \  \ C \,) 
\quad \{\wedge \ \ D \ \   (\vee \ \ A  \ \ \overline{B} \,) \,\}  \,\}$
 \qed
\end{example}

\begin{proposition}
   An HNF logic program is an HNF formula.
 \end{proposition}
 
\begin{niceproof}
Clearly a rule $ (\vee \  \neg \Pi^+  \ \Phi )$ satisfies Definition \ref{theorem:visual} and hence 
 so does a conjunction of them, i.e. an HNF logic program.
\end{niceproof}

\begin{lemma} \label{lemm:Complexity-Program}   Let $ \mathcal{P} $ be a  proposition set,  
 Lp   an HNF logic  program and $ \varphi $  any arbitrary NNF. 
 Deciding whether $\mathcal{P}  \wedge Lp \models \varphi $   is polynomial.
\end{lemma}

 \begin{niceproof}  By Lemma  \ref{lem:polytime},  one  can polynomially check  
whether $ \mathcal{P}  \wedge Lp$ is satisfiable. If  so, by applying UR$_{\mbox{nc}}$ one can polynomially determine
 the positive 
literals that follow logically from 
$ \mathcal{P}  \wedge Lp$, i.e. its minimal model.  
Finally, one can 
 also polynomially check  whether, for such minimal model,  $ \varphi $   evaluates to  1
 and so whether $\mathcal{P}  \wedge Lp \models \varphi $.
\end{niceproof}

\noindent Given Corollary \ref{cor:firstcorprimer}, each HNF
program is equivalent to a clausal definite program, thus one can say
that the HNF programs are the nested definite programs. On the other
side,
  Example \ref{ex:NCrule} visualizes  the potentiality  
of  $ \mathbb{{H}_{NF}}$  to  enrich     declarative rules  in 
   logic programming.
As mentioned previously, we think that future work will allow
to find data-structure to linearly test the satisfiability of $ \mathbb{{H}_{NF}}$.
 This would imply that the complexity of propositional HNF logic programs is linear, 
that is,    expressiveness  can be significantly  augmented    
while keeping the same linear complexity  
of clausal logic programming \cite{DantsinEGV01,BryEEFGLLPW07,Warren18}. 
Besides, HNF logic programs are much smaller than their equivalent Horn  programs.

\section{Additional    Properties of $\mathbb{H_{NC}}$}
\label{sec:semanticalsyntactical}

 This section proves three additional properties of $\mathbb{H_{NC}}$. 

\subsection{Linear Recognition of $\mathbb{H_{NC}}$} \label{subsec_SynProp}

 We   present a recognition algorithm,  {\bf HNF}, to determine whether an  NNF  is
HNF and  also        prove   its correctness and linearity.
Checking whether an   NC 
$\varphi$ is  HNC is performed by first translating $\varphi$   into 
an equivalent NNF $\varphi'$, and then, by applying   {\bf HNF} to  $\varphi'$.
That translation can be achieved in  linear-time by
  merely pushing in the negations.




\vspace{.15cm} 
\noindent $ \bullet $ {\bf  {\bf HNF}'s Principle.}
 We call   disjunctions and conjunctions  of literals, e.g. $(\vee \ \overline{A}  \ B)$, flat-formulas.  
%
   {\bf HNF} uses two  fresh literals, 
    $\ell^+$ and $\ell^-$,   treating them  
 as  positive and negative, respectively, and
  replaces           flat-formulas of the input with $\ell^+$ or $\ell^-$ as described below:
  
  \vspace{.1cm}
   $(1)$ The strategy to bottom-up   replace   
        flat-formulas is as follows. In a given step, {\bf HNF} replaces   the  flat-formulas
  with   $\ell^+$ or $\ell^-$. So,
     sub-formulas  containing  only literals and flat-formulas 
     are flattened and
    so  recursively replaced in the next step.
    
    \vspace{.1cm}
     $(2)$  Below in {(a)}-{(e)},  we enumerate
the five possible cases encountered  to replace   a given flat-formula $\phi$   with 
 either $\ell^+$ or $\ell^-$.  
     $pos(\phi)$    is the number of positive   literals of $\phi$. If case {(a)} holds,
    then  {\bf HNF} halts and signals that the input is {\em non-HNF.} 
      
\vspace{-.1cm}       
\begin{itemize}  
\item  [$\bullet$] If  $\phi$ is a   clause,    then
{\bf HNF}  operates as follows:

\vspace{-.3cm}
\begin{enumerate}

\item [(a)]  $pos(\phi) > 1$ \ $\longrightarrow$  \  halts  and  returns {\em non-HNF}. 

\item  [(b)]  $pos(\phi) = 1$   \  $\longrightarrow$ \ replaces    $\phi$    with $\ell^+$. 

\item  [(c)] $pos(\phi) = 0$ \  $\longrightarrow$ \   replaces    $\phi$   with  $\ell^-$. 
\end{enumerate}

\vspace{-.3cm}
\item  [$\bullet$] If $\phi$ is a conjunction,  then  {\bf HNF}  acts as follows:

\vspace{-.3cm}
\begin{enumerate}
\item  [  (d)]  $pos(\phi) >  0$  \    $\longrightarrow$ \   replaces    $\phi$    with $\ell^+$.

\item  [  (e)]  $pos(\phi) =  0$ \ $\longrightarrow$ \  replaces    $\phi$    with $\ell^-$.

\end{enumerate}
\end{itemize}

 $(3)$  {\bf HNF}  returns three possible labels either
 {\em non-HNF}, $\ell^-$ or $\ell^+$  
  whose  meanings are    $\varphi \notin \mathbb{H_{NF}}$, 
     $\varphi \in   \mathbb{N_G}$, and  
    $\varphi \in \mathbb{H_{NF}} / \mathbb{N_G}$
(non-negative HNF), respectively. 
Hence,
the correctness of {\bf HNF} amounts to:
 {\bf HNF}($ \varphi $) returns   "non-HNF" iff   $ \varphi $ is not HNF.


\vspace{.2cm}
\noindent $\bullet$ {\bf Pseudo-Code.} The required data-structure 
is:
%
%
     An NNF  
  is   implemented by associating to each sub-formula   $\varphi$: 
   (1) a pointer  $[\varphi]$
 that points to  the   address  where  
 the  linked structure of symbols constituting $\varphi$ begins, i.e. 
to the root of  DAG D$_\varphi $. 
D$_\varphi $ is stored only once and
each $ \varphi $-occurrence in the   NNF  is   replaced with   $ [\varphi] $; and 
 (2) an integer $pos(\varphi)$      
 that, when  $\varphi$ is flattened,       is set to the  number of positive 
literals in $\varphi$.  The pseudo-code of {\bf HNF}   is given above;  the right  column
   indicates     cases {(a)}-{(e)} in the 
 above    {\bf HNF}'s principle   corresponding to
 the   line in question.

\vspace{.2cm}
{\bf \,HNF$( \,[\varphi]\,)$}.

\vspace{.1cm}
\hspace{.5cm} {\bf If} \ \  $[\varphi]=(\vee \ \ell_1 \,  \,\ldots  \ell_{i} \ldots \,\ell_k)$ \   {\bf then do:} 

\vspace{.1cm}
\hspace{1.5cm}   $pos( \,\varphi\,) \leftarrow$   number of  positive-literals in $\varphi$.

\vspace{.1cm}
\hspace{1.5cm}  {\bf if} \ \ $pos(\,\varphi\,) > 1$ {\bf then}  halt and return {\em non-HNF}. \ \ \,:   {(a)}

\vspace{.1cm} 
\hspace{1.5cm}  {\bf if}  \ \   $pos(\,\varphi\,)=1$ {\bf then}  return($\ell^+$). \hspace{2.7cm} :   {(b)}

\vspace{.1cm}
\hspace{1.5cm}  {\bf if}  \ \  $pos(\,\varphi\,)=0$ {\bf then} return($\ell^-$). \hspace{2.7cm} :   {(c)}

\vspace{.1cm}
\hspace{.5cm} {\bf If} \ \  $[\varphi]=\{\wedge \ \ell_1 \,\ldots \,\ell_{i} \ldots \,\ell_k\}$ \   {\bf then do:} \   

\vspace{.1cm}
\hspace{1.5cm}    $pos( \,\varphi\,) \leftarrow$ number of positive-literals in $\varphi$.

\vspace{.1cm}
\hspace{1.5cm} {\bf if} \ \ \,$pos(\,\varphi\,) > 0$ {\bf then} return($\ell^+$).\hspace{2.8cm} :  {(d)}

\vspace{.1cm}
\hspace{1.5cm}  {\bf if} \ \ \,$pos(\,\varphi\,)=0$ {\bf then} return($\ell^-$). \hspace{2.65cm} :   {(e)}

\vspace{.15cm}
\hspace{.5cm} {\bf  else} \  for   $1 \leq i \leq k, 	\,\varphi_i \neq \ell_i$ \,{\bf do:} \ $[\varphi_i] \leftarrow $  {\bf HNF}(\,$[\varphi_i]$\,). 

\vspace{.15cm}
\hspace{1.5cm}    {\bf Return HNF}(\,$[\varphi]$\,).

\noindent

\begin{example}  Next, we run {\bf HNF} on $\varphi  $ below.
Its flat-formulas   are  $\phi_1=(\vee \  \overline{A}  \ E)$,   
$$\varphi=\{\wedge \ \ \overline{C} \  \ (\vee \ \ \overline{A}  \ E \,) 
 \  \ (\vee  \ \ (\vee \  \    
\{\wedge  \  \ \overline{F} \  \overline{C} \, \} 
\ \  \overline{E} \,) \ \  \{\wedge \ \  A \  B \, \}  \, ) \,\}.$$

\noindent 
$\phi_2=\{\wedge  \   \overline{F} \  \overline{C} \}$   
     and  $\phi_3=\{\wedge \  A \  B \}$, and their associated replacements  are:         
     \begin{enumerate}
      \item     $pos(\phi_1)=1$ \  $\longrightarrow$  \ case  (b) \ $\longrightarrow$  \  $\phi_1$ is replaced with $\ell^+$.
  
  \vspace{-.25cm}   
 \item    $pos(\phi_2)=0$ \ $\longrightarrow$ \  case    (e) \   \,$\longrightarrow$ \   $\phi_2$ is replaced with $\ell^-$.
    
     \vspace{-.25cm}    
   \item    $pos(\phi_3)=2$ \  $\longrightarrow$ \  case   (d) \  $\longrightarrow$  \   $\phi_3$  is replaced  with $\ell^+$.
     \end{enumerate}
     
 --   After   the   replacements,  one obtains: \ 
$\{\wedge \  \  \overline{C} \ \   \ell^+  
\  \ (\vee \ \   (\vee \  \  \ell^- \ \overline{E} \,) \  \ \ell^+    \,) \,\}.$ 
 
 \vspace{.15cm}
-- Replacing  its  flat-formula      $(\vee \  \ell^- \  \overline{E} \,)$   with   $\ell^-$ yields:               
$\{\wedge \   \ \ \overline{C} \ \  \ell^+ \  \ (\vee  \  \ \ell^- \  \ell^+  \,) \,\}.$

 \vspace{.15cm}
-- Replacing   its   flat-formula     $(\vee  \  \ell^- \  \ell^+    \,)$   with   $\ell^+$ yields:  
 $\{\wedge \  \  \overline{C} \ \  \ell^+ \  \  \ell^+ \,\}.$
 
 \vspace{.15cm}
--  Here, {\bf HNF}   returns $\ell^+$, indicating
   that $ \varphi $   is   non-negative HNF. \qed
     
\end{example}


\begin{example} We now run {\bf HNF}  on the   formula:
$\{\wedge \ \ \overline{C} \ \ (\vee  \ \ \{\wedge \ \  A \ \ \overline{C} \,\} \ \ E \,) \,\}.$

\vspace{.2cm}
-- Replacing $\{\wedge  \ \ A \ \  \overline{C} \,\}$  with  $\ell^+$ leads to
 $\{\wedge \   \  \overline{C} \ \  (\vee  \ \  \ell^+  \ \ E   \,) \,\}.$

 \vspace{.1cm}
--   As $pos(\vee \  \ell^+ \   E ) > 1$,  condition    (a) holds and      {\bf HNF}
  halts  
   returning {\em non-HNF}. \qed
\end{example}




\begin{theorem} \label{Th:correctness} Correctness of {\bf HNF}: \ $\varphi$  is  non-HNF  
iff {\bf \,HNF}{\em ($\varphi$)} returns  non-HNF.
\end{theorem}

\begin{niceproof} We denote  $nn(\phi_{init})$  the number of non-negative disjuncts
of any initial disjunctive sub-formula $ \phi_{init} $ of  the input $ \varphi $. 

$\bullet  $ Steps  (b)  to  (e).  A flat-formula $\phi$ is
  replaced with  $\ell^+$   iff  $\phi$  has some positive literal. Assume that $\phi'$ is the father of
  $\phi$. As both $\ell^+$ and $\phi$ are non-negative, 
  $nn(\phi')=nn(\phi)$. Since
sub-formulas are recursively flattened,  all
initial   $nn(\phi_{init})$ are recursively preserved. 
%
%
Hence if $\phi_{flat}$ is the initial   $\phi_{init}$ after having 
been flattened,  then  $pos(\phi_{flat})=nn(\phi_{init})$.
%

$ \bullet $ Step (a). Assume that      (a)   sometime  holds.   If  {\bf HNF}    
       halts  and returns   {\em non-HNF}, then some $pos(\phi_{flat}) > 1$ is encountered. But   
   $pos(\phi_{flat})=nn(\phi_{init})$ and  $pos(\phi_{flat}) > 1$   entail that  
 $\exists \phi_{init}, \ nn(\phi_{init}) > 1$, and so,   by 
       Definition \ref{theorem:visual},  the input  $\varphi$ is not HNF. 
        Hence, if  {\bf HNF} returns {\em non-HNF},    $ \varphi $ is non-HNF.
On the contrary,   supposing that the hypothesis     (a)  is never met,   then by following 
     a reasoning dual to the previous, one  obtains that if 
     {\bf HNF} returns $\ell^+$ or $\ell^-$, then $ \varphi $ is HNF.    
\end{niceproof}
 

\begin{theorem} Linearity of {\bf HNF}: \  {\bf HNF$(\varphi)$} ends in time 
$O(size(\varphi))$.
\end{theorem}

\begin{niceproof}   The running time of {\bf HNF} is the sum    of   the times  for
 {\em building} and   {\em scanning}
  (applying cases (a)-(e)) the DAG
 associated  to $\varphi$.
  {\em Building:}    pointers     
     are initialized and  counters    allocated;   both in constant time; 
   so building is   linear. 
 {\em   Scanning}:  
$(i)$ arcs are traversed at most once upwards and  once downwards.
  $(ii)$   when  a  sub-formula   is flattened,  
    {\bf HNF} counts its   positive literals     
    and set $pos(\phi)$  accordingly, 
      which   is done   at most  once  and takes   
       $O(k)$ time,   $k$ being the arity  of $\phi$; so
       $(ii)$   is linear.  
 $(iii)$ replacing   a   flat-formula  
   with $\ell^+$  or $\ell^-$ is done at most  once and in  constant time. Hence,
   the overall cost   is linear.
\end{niceproof}

\subsection{ Succinctness and Syntactical Richeness}  \label{sub-sec:preliminaries}

Succinctness  was  introduced in \cite{GogicKPS95} and is   given in Definition \ref{def:succinctness}. 
It measures spatial 
efficiency of formula classes  for knowledge modeling and has been used   in many works.

\begin{definition} \label{def:succinctness}  Let $\mathbb{X}$ and $\mathbb{Y}$ two  formula classes.
 $\mathbb{X}$   is at least as    succinct as $\mathbb{Y}$, noted $\mathbb{X} \leq \mathbb{Y}$, iff 
for every $\varphi \in \mathbb{Y}$ there exists $\varphi' \in \mathbb{X}$  
    s.t. $ \varphi\equiv \varphi' $ and
the size of $ \varphi' $ is polynomially bounded in that of 
$\varphi$.
 $\mathbb{X}$ is    strictly more  succinct than $\mathbb{Y}$ iff \,
$\mathbb{X} \leq \mathbb{Y}$ and $\mathbb{Y} \nleq \mathbb{X}$.
\end{definition}

\begin{proposition}    $\mathbb{H_{NC}}$ is  strictly more succinct than   $\mathbb{H}$.
\end{proposition}

\begin{niceproof}  Clearly $\mathbb{H_{NC}} \leq \mathbb{H}$  because 
$\mathbb{H_{NC}} \supseteq \mathbb{H}$.
To prove that $\mathbb{H} \nleq  \mathbb{H_{NC}}$, 
it is enough to   take
 any DNF   $\varphi \in \mathbb{H_{C}}$ with $k$ conjuncts
and $n$ literals per conjunct.  Thus $cl(\varphi) \equiv \varphi \in \mathbb{H}$ and the size of $cl(\varphi)$ is  $k^n$. 
\end{niceproof}


We now deal with syntactical richness and
consider  that   class  $\mathbb{X}$ is richer than  
 $\mathbb{Y}$ simply if $\mathbb{X}$  subsumes $\mathbb{Y}$.
As formulas   express interrelations among variables,   
richer classes having more formulas supply     more syntactical options 
to express  interrelations among variables and
 so  have  superior syntactical capabilities for knowledge modeling.
%
Clearly   $\mathbb{H_{NC}}$  is richer than     $\mathbb{H}$, but    
  we check  that
  it is even exponentially richer.

\begin{definition} \label{def:exp-rich}  Let    $\mathbb{X}$ and
    $\mathbb{Y}$ be two   formula classes.      $\mathbb{X}$   
    is  exponentially richer than 
     $\mathbb{Y}$    iff:   
\begin{enumerate}
\vspace{-.1cm}
\item [$($i$)$] $\mathbb{X} \supset \mathbb{Y}$.
\vspace{-.2cm}
\item [$($ii$)$]     $\forall \varphi' \in 
\mathbb{Y}$ there are exponentially many 
  $\varphi \in \mathbb{X}$ s.t. $\varphi \equiv \varphi'$.
\end{enumerate}  
\end{definition}

\begin{proposition} \label{lem:CNFexponent}     $\mathbb{H_{NC}}$ is  exponentially richer than
      $\mathbb{H}$. 
\end{proposition} 
 
 The proof is straightforward.
 Note, however, that the previous proposition 
 does not imply that $\mathbb{H_{NF}}$ is exponentially richer than $\mathbb{H}$.

 


\section{Related Work} \label{sec:relatedwork}

  Given that  $ \mathbb{H_{NC}} $ results from merging  
   the NC   and   Horn formulas,   we
review   the main literature relative to both of kind of formulas.

\subsection{Non-Clausal Reasoning} \label{subsec:non-clausal}

   
\noindent {\bf Logic Programming.} This area has been discussed in Subsection \ref{sub:applic-log-prorm}.

\vspace{.1cm}
\noindent {\bf Satisfiability.} Given
 the    growing  efficiency
of clausal DPLL  solvers, enhancing  NC reasoning   tended 
toward translating  NCs  
into equisatisfiable clausal ones   via Tseitin-like  \cite{Tseitin83} translators,   
and  then, applying   DPLL   solvers. Yet,   the  benefits of translations 
were questioned    
\cite{ThiffaultBacchusWalsh04,NavarroVoronkov05,DrechslerJunttilaNiemela09}, 
which  advocated  both, developing   approaches  computing
  original NCs and incorporating    structural information 
in  the search.
%
On the other side, a number of   approaches for NC satisfiability exist: NC Resolution \cite{Murray82}, NC DPLL  \cite{Otten97,GiunchigliaSebastiani99,ThiffaultBacchusWalsh04,JainClarke09},
path dissolution  and anti-links \cite{RameshBeckertHahnleMurray97},  
 connection calculi \cite{Otten11},  regular, connected tableaux  
\cite{HahnleMurrayRosenthal04},  NC
tableaux-like with
 reductions \cite{GutierrezGuzmanMartinezOjedaValverde02},
SLD-resolution \cite{Stachniak01A}  and
  general matings \cite{JainBartzisClarke06}. 
  The  experimental results    showed  the relative good performance of  DPLL.   
%
 Although several NC DPLL solvers  have been implemented, NC unit resolution  had
  not been published.  On the other side, 
  a rule called NC unit-dissolution  is introduced in \cite{MurrayR94c}, but
   this is done within the framework of other general rule
   (Dissolution) and it is not formalized.
 Our  work led to formally define  NC unit-resolution for the
 first time,
 prove its  completeness for $\mathbb{H_{NC}}$ 
 and it allowed 
 to demonstrate the polynomiality  of $\mathbb{H_{NC}}$.  

\vspace{.15cm}
\noindent {\bf Incomplete Solvers.} 
For clausal problems, GSAT 
\cite{SelmanLM92} demonstrated  that an incomplete   solver can solve many
hard problems much more efficiently than  complete   solvers. 
Several authors have   generalized
these techniques to NCs. Sebastiani \cite{Sebastiani94} suggested how to modify
GSAT to be applied to NCs.
Kautz et al. \cite{KautzSM97} introduced DAGSat for clausal formulas,
but allowed handling of NCs without an increase in size. Later,
Stachniak \cite{Stachniak02} introduced polWSAT an evolution of WalkSAT \cite{SelmanKC94} to handle
NCs. The authors in \cite{MuhammadS06}  present a new way of expressing the scoring function for evaluating  
partial interpretations and their solver  outperforms previous  complete
and incomplete NC solvers. Yet, as HNCs are 
polynomial,    designing
 incomplete  NC solvers  for them has limited interest.
  UR$_{\mbox{nc}}$ could be
 relevant 
to develop  incomplete NC-SAT solvers.

\vspace{.15cm}
\noindent {\bf Quantified Boolean Formulas.} 
An extension of  the SAT problem 
 increasingly  present in research and applications,
 is   featuring universal and existential quantifiers for     boolean variables giving rise   
 to the QBFs. In particular, developing efficient 
  solvers for  QBFs in NC form is a  concern of contemporary research    
 as   evidenced  by the    reported  performances,    e.g.
    \cite{LonsingB08, EglySW09,BalabanovJ12}.
    However, to the best of our knowledge, no results presented here  concerning
    $\mathbb{H_{NC}}$ and UR$_{\mbox{nc}}$  have been initialized
     for QBFs in NC.

\vspace{.15cm} 
\noindent {\bf Max-SAT.} 
  Some reasoning  
problems have been intensively studied 
for clausal but  hardly for NC, and    sometimes,  
even  no truly NC definition   exists. 
      MaxSAT is an example.
The    reached  efficiency
for   MaxSAT  \cite{ChuMinLiManya09} 
 contrasts with the 
   nonexistence of a proper definition for
\mbox{NC-MaxSAT}.
%
 A recent work  \cite{LiMS19}  
defines   NC-MaxSAT  as the problem of, given
$k$ NCs
\{$\varphi_1, \, \varphi_2, \ldots, \varphi_k$\}, 
determining the greatest   $i$,  $i \leq k$,     of   
     simultaneously satisfiable formulas. 
In  such definition the only envisaged formula inter-dependence    is
   the one derived from   the   shared literals  which  is, however, 
   rather   clausal-like.
The proper  NC inter-dependence 
      is the one  derived from the  connexions among 
       formulas and   sub-formulas.  
%
 Applications  of clausal Unit Propagation    include its use 
for improving the lower-bound during the solving of 
 MaxSAT \cite{LiManyaPlanes07}. Thus 
 UR$_{\mbox{nc}}$ 
   seems essential to
   compute efficiently   \mbox{NC-MaxSAT}.

\vspace{.15cm}
\noindent {\bf  Theorem Proving.}  Automated reasoning in NC  and FOL has been an  
active  research field since the  1960s and    is  still the object of  considerable research 
   and  progress,  e.g.  \cite{HahnleMurrayRosenthal04,DBLP:journals/entcs/HolenJW09,Otten11,FarberKaliszyk19}.   
  On the other side, DPLL  for clausal FOL has already been developed \cite{BaumgartnerT08,Baumgartner14}.
 Extending NC FOL with equality logic is a former \cite{BachmairG92} as well a recent   \cite{OliverO20} concern.
  Although our approach has been initially oriented towards propositional logic,  it is extensible  to FOL. Concretely, we  plan to propose a FOL NC DPLL by
  lifting UR$_{\mbox{nc}}$  to FOL (Future Work), which
  means, moreover, to generalize to  clausal  approach   in \cite{BaumgartnerT08,Baumgartner14}.

\vspace{.15cm}
\noindent {\bf  Resolution.}   In the seminal  work by N. Murray  \cite{Murray82},    
 NC Resolution  for propositional and FOL
  was defined  in a combined functional and classical manner and has been applied
  to different  contexts, e.g. \cite{BachmairG92,Stachniak01A}. Each proper
  NC Resolution   is followed by some functional-logical simplifications of the formula.
  The completeness of NC Resolution {\em under a linear restriction}   was proved 
    by the authors in  \cite{ HahnleMurrayRosenthal04}.
  Although NC Resolution was specified in 
  1982, paradoxically  
no NC Unit-Resolution version  had   been  published.  NC Hyper-resolution  
had  not been defined  either to   our knowledge. 
The available definition of NC  Resolution suffers
form some drawbacks e.g. not clearly defining the potential
resolution steps or requiring complex completeness proofs
 \cite{ HahnleMurrayRosenthal04}. We plan to present a new
  definition of NC Resolution (see future work).

\vspace{.15cm}
\noindent {\bf Knowledge Compilation.} It consists in translating an original formula    
  into another equivalent  one  during the off-line phase.
Whereas  the original and target formulas are semantically equivalent,  the target class
  present  desired syntactical conditions 
 permitting computing  the clausal entailment test   polynomially. 
 This way, the on-line exponential complexity,  usually present during query-answering, is  confined to  the off-line  phase  and 
   later amortized over all
 on-line queries answered  polynomially.  
 NNF   sub-classes, such as DNNF,  d-NNF
or the well-known BDD and OBDD \cite{DarwicheM02},
 were introduced as target ones  by A.  Darwiche 
\cite{Darwiche99,DBLP:journals/jacm/Darwiche01}.
   $ \mathbb{H_{NC}} $   is
polynomial for satisfiability checking, i.e. for clausal entailment, but 
    $ \mathbb{H_{NC}} $ is incomplete
 since not all NNFs are of course equivalent to, and so  representable by,  
an  HNC.  So,
 $ \mathbb{H_{NC}} $ does not qualify as a target  class.
 Yet in \cite{FargierM08},
 the   class  formed by the disjunctions of Horn   formulas, or  $ \mathbb{H} $-[$\vee$],  is introduced as a target class. learly, the class   of disjunctions of HNCs, or $ \mathbb{H_{NC}} $-[$\vee$],  subsumes  $ \mathbb{H} $-[$\vee$] and is strictly 
 more succinct, 
 likewise $ \mathbb{H_{NC}} $ is strictly more succinct than  $ \mathbb{H} $. 
 Towards using $ \mathbb{H_{NC}} $-[$\vee$] as a target class,
 the next  questions should be analyzed: how to compile knowledge into $ \mathbb{H_{NC}} $-[$\vee$]; 
  which is
  its  complexity for  consistency,   clausal entailment,
  implicants, etc. \cite{DarwicheM02}; and finally, 
  its succinctness should be compared with existing target classes.

\subsection{Horn Formulas} \label{subsec:Horn-reason}

Due to the ubiquity of  Horn formulas
in  knowledge representation and automated reasoning, and
   by space reasons, 
our related work  is unavoidably  incomplete.
It  focuses first on their presence   in knowledge representation
   and then
 in automated reasoning.
 



\subsubsection{Knowlegde Representation.} \label{subsec:Horn-Super-classes}


\noindent {\em \bf  Propositional logic.} As alluded to in the introduction,
finding polynomial clausal  super-classes of  the  Horn  class has been
a key issue on increasing  clausal efficiency   since 
    1978 \cite{Lewis78}.   Most  of such extensions
    are presented below chronologically along with
       their   relationships.
Our research   has allowed   to   lift  the Horn formulas to  NC and  
       prove  that tractable reasoning also exists   in NC, but we think that this is just a first
        outcome and that some Horn extensions pointed out below can also be  
       NC lifted. For instance, one can conceive
       the renomable-HNC formulas as those NCs whose  clausal 
       form is renomable-Horn.

\vspace{.15cm}
\noindent   {\it $\bullet \ \mathsf{Unit \ Propagation}$}.\ The  so-called {\it Unit Propagation} was proved    complete
for Horn-SAT in \cite{HenschenWos74,GalloUrbani89,Minoux88}. Further publications 
 \cite{ItaiMakowsky82,DowlingGallier84,Scutella90}   get this universally understood.  
The linearity of  Horn   algorithms   \cite{DowlingGallier84,ItaiMakowsky87,Minoux88, Scutella90,GhallabEscalada91,DuboisAndreBoufkhadCarlier96} gave rise to  efficient implementation of 
 {\it Unit Propagation}.    
Deciding if a model of a  Horn  formula 
is unique  is also   linear   \cite{Petrolani93}. 

\vspace{.1cm} 
\noindent  {\it $\bullet \ \mathsf{Hidden \ Horn}$}.\  A given  formula is 
hidden Horn  \cite{Lewis78,Aspvall80} if a change in the polarity of some 
variables leads to an equivalent  Horn  formula. 
  Recognizing  hidden Horn formulas
is linear \cite{Aspvall80,ChandruCoullardHammerMontanezSun90} and since   Horn-SAT  is linear,
  so is SAT-checking    them.

\vspace{.1cm}
\noindent   {\em $\bullet \ \mathsf{S_0, \ Hierarchy \ \varphi_K}$}. \  Another  Horn  extension  was proposed in  
\cite{YamasakiDoshita83}, called  $S_0$, and later  Generalized  Horn  in \cite{ChandruCoullardHammerMontanezSun90},
and  successively broadened in \cite{ArvindBiswas87,GalloScutella88}.
 Recognizing   and SAT-checking $S_0$ formulas   
are  quadratic \cite{ArvindBiswas87,GalloScutella88}. Reference \cite{GalloScutella88}
defines recursively a hierarchy 
$\Sigma_0 \subseteq \Sigma_1  \ldots \subseteq \Sigma_k  \ldots $  over
the non-clausal classes $\Sigma_k$,  where $\Sigma_0$ are the  Horn  formulas and $\Sigma_1$ is $S_0$. This hierarchy was strengthened in \cite{Petrolani96} whose classes $\Pi_k$
 subsume the classes $\Sigma_k$. SAT-checking  
 $\Sigma_k$  and  $\Pi_k$ formulas 
      is polynomial  but recognizing         the  renamings of
$\Sigma_k$ formulas,  $k > 1$, is $\mathcal{NP}$-complete 
\cite{EiterKilpellinenMannila95}.

\vspace{.1cm}
\noindent   $\bullet \ \mathsf{Q\cdot Horn}$.   The   Q-Horn  formulas \cite{BorosCramaHammer90}  subsume the  
    {\em  Horn},  {\em hidden Horn}  and the {\em Quadratic} (2-CNF) formulas.    {\it Q-Horn}
formulas are  recognized  and SAT-checked  
linearly  \cite{BorosCramaHammer90,BorosCramaHammerSaks94}.

\vspace{.1cm} 
 \noindent     $\bullet \ \mathsf{Extended\  Horn}$.   The {\it   extended-Horn} formulas were introduced 
in \cite{ChandruHooker91}  based on a theorem in
\cite{Chandrasekaran84}. Their definition relies on {\em Unit Propagation}
and matrix multiplication.  {\it Hidden extended-Horn formulas} are
defined   in the same way as  {\em hidden Horn}   from {\em Horn.}
 SAT-checking {\it extended-Horn} formulas   is linear; however,  no polynomial recognition
of the {\it extended-Horn} formulas is  known.

\vspace{.1cm}
\noindent {\it  $\bullet \ \mathsf{Hierarchy \ \Omega}$}.\ The {\it  hierarchy $\Omega$ 
of classes} proposed in \cite{DalalEtherington92} 
contains the classes  {\em Horn, reverse-Horn} (clauses have at most one negative literal)
and {\em hidden-Horn.}
 The classes in $\Omega$ are recognized  and SAT-checked polynomially.

\vspace{.1cm}
\noindent    $\bullet \ \mathsf{SLUR}$. \ The definition of the {\it  SLUR (Single Lookahead Unit Resolution)} formulas \cite{SchlipfAnnexsteinFrancoSwaminathan95,BalyoGurskyKuceraVlcek12}  is based on an algorithm 
rather than on properties of formulas. 
SAT-checking a formula $\varphi$  is divided in two phases. First unsatisfiability is sought 
by applying {\em Unit Propagation} (UP). If such seeking fails, 
then $\varphi$ is assumed to be satisfiable and
a satisfying assignment is guessed  using UP-lookahead to avoid obviously false assignments. 
The class {\em SLUR} contains those formulas  where the algorithm succeeds
in the first phase or finds a satisfying assignment in the second phase.
The class {\em SLUR}
subsumes the
{\it hidden-Horn, 
extended-Horn}  and other
classes \cite{SwaminathanWagner95}.  
 {\it SLUR and Q-Horn} 
are incomparable \cite{FrancoVanGelder03}.  The {\it SLUR formulas} are efficiently 
SAT-checked but  recognizing them 
   is  co{\bf NP}-complete \cite{CepekKuceraVlcek11}.

\vspace{.1cm}
\noindent {\it $\bullet \ \mathsf{Quad}$}.\ The  class {\it Quad}  \cite{Dalal96} includes the {\em Horn} and 
the {\em Quadratic} (2-CNF)
formulas.
 {\it Quad} is not subsumed  by the hierarchy $\Omega$, and  {\it Q-Horn} 
and {\it Quad} are incomparable. Recognizing  and SAT-checking {\it Quad} 
formulas take quadratic time.

\vspace{.1cm} 
\noindent {\it $\bullet \ \mathsf{Autarkies}$}.  If a partial assignment
  satisfies all those clausewe 
  thinks of $\varphi$
 affected by
it, i.e., no new clauses are created,  the resulting formula is an  {\it autarky,}  e.g. the simplest autarkies are the pure literals. Therefore, $\varphi$ is satisfiable iff its autarky is. The  class of {\it  autarkies} 
    include    {\it Q-Horn} and is
incomparable with the class {\em SLUR} \cite{vanMaaren00}. 
  Finding  the specific  linear {\em autarkies}  is polynomial \cite{Kullmann00}
 and there    exist partial {\em autarkies } that applied
repeatedly result in a unique, autarky-free
formula.

\vspace{.15cm}
\noindent {\it $\bullet \ \mathsf{Matched}$}.  The  {\it matched} formulas were analyzed 
initially in \cite{Tovey84} in order to provide a benchmark for
testing some previously published  formulas but  were considered rather useless.
However, although the class {\it matched} is incomparable with  {\it Q-Horn} and {\it SLUR},   
w.r.t. frequency of occurrence on
random formulas,
 reference \cite{FrancoVanGelder03}  verified
that  {\it matched}  is far more common than both those
classes together. {\it Matched} formulas are always satisfiable
  and trivially
SAT-checkable.

\vspace{.1cm} 
\noindent {\it $\bullet \ \mathsf{Minimally \ Unsatisfiable}$}. \  Several classes of  unsatisfiable formulas are
identified in \cite{Kullmann03}. A  formula is {\it minimally unsatisfiable} if it is unsatisfiable   
and removing any clause results in a satisfiable formula  \cite{BuningKullmann09}.  
{\it Minimally unsatisfiable formulas} are SAT-checked  polynomially.

\vspace{.1cm} 
\noindent {\em $\bullet \ \mathsf{Various \ results}$}  \cite{CepekKucera05}. \ 
 The next  results are provided in \cite{CepekKucera05}:  (i) none of the classes $S_0$, 
{\it  Q-Horn}  and {\em extended Horn} 
 subsumes any other;  (ii)     a  new class
is defined that subsumes $S_0$; and (iii) originated in that new class,   a 
 hierarchy of classes is proposed that generalizes 
 $\Omega$  (see   hierarchy $\Omega$ above) 
and  the classes $S_0$ and {\em Q-Horn.}  Recognizing 
and SAT-checking the 
newly described, hierarchical classes 
are both polynomial.

\vspace{.1cm} 
\noindent {\it $\bullet \ \mathsf{CC\cdot balanced}$}. \ The  {\it CC-balanced} formulas have been studied in
 \cite{Confortietall06} and the motivation for this class is   when
%
Linear Programming relaxations of SAT-checking have non-integer solutions.
Recognizing  and SAT-checking {\it CC-balanced} formulas  are 
linear.

\vspace{.1cm} 
\noindent  {\it  $\bullet \ \mathsf{UP\cdot Horn}$}. \
The   {\it  UP-Horn} formulas \cite{FourdrinoyGrgoireMazureSais07} include
 the   Horn   and the non-Horn formulas  resulting in  Horn  formulas
after applying the   {\em Unit Propagation}. Thus,  recognizing and SAT-checking {\it UP-Horn} formulas are polynomial. 

\vspace{.1cm} 
\noindent {\em $\bullet \ \mathsf{SLUR \ hierarchies}$}.    
{\em SLUR}($i$) \cite{CepekKuceraVlcek11} is a  hierarchy   on top of the class  SLUR.
 Level $i$-th consists in 
making the {\em SLUR} algorithm to consider simultaneously
$i$ literals and their corresponding $2^i$ assignments. {\em SLUR}*($i$)  \cite{BalyoGurskyKuceraVlcek12} 
is a  hierarchy more general than {\em SLUR}($i$) and
 relies on allowing DPLL  to backtrack at most $i$ levels. The class $SLUR_K$ is introduced
 in \cite{GwynneKullmann14} and proved that it strongly subsumes previous  hierarchies {\em SLUR}($i$) and {\em SLUR}*($i$).
 The authors in \cite{GwynneKullmann14} develop  another polynomial SAT 
 class $UC_K$ based
 on the class {\em Unit-refutation Complete},  $UC$  \cite{delVal94}, and  prove a major result, concretely that $SLUR_K=UC_K$.

\vspace{.1cm}  
\noindent {\em $\bullet \ \mathsf{Various \ results}$} \cite{AlSaediFourdrinoyGregoireMazureSais17}. \  Reference  \cite{AlSaediFourdrinoyGregoireMazureSais17} 
presents some new polynomial
 fragments based on {\em Unit Propagation}: 
 (i)  the polynomial class  in \cite{Tovey84}, formed by formulas whose
 variables may appear at most twice, is extended; 
(ii)   some series of benchmarks from
the DIMACS repository and from  SAT competitions are shown to belong
to {\em UP-Horn}; and
(iii) some set relations between the classes {\em Quad} 
 and  {\em UP-Horn}  are established.

\vspace{.3cm} 
\noindent {\em \bf  Beyond Propositional Logic.} 
  Horn  formulas are used in a large variety of artificial 
 intelligence logics. Hence, we make  reference just to  some logics somewhat close 
 to propositional logic and  whose NC lifting    is discussed in Future Work.


\vspace{.15cm}
\noindent $ \bullet $    $\mathsf{  Regular \ Many \cdot Valued}$ and $\mathsf{Possibilistic \ Logics.}$ The regular many-valued Horn formulas \cite{Hahnle01} and the
  possibilistic Horn formulas \cite{Lang00, DuboisP94,  DuboisP14} have been  lifted   to NC
 in 
 \cite{Imaz21b,Imaz2021c}.

\vspace{.15cm}  
\noindent $ \bullet $ {\em $\mathsf{Linear \ Arithmetic \ Logic.}$} 
  In     linear arithmetic  logic  \cite{AudemardBCKS02b,AudemardBCKS02}, 
    an atom is a   disequality,  e.g. $4x_1 + x_3 \neq 3$, 
  or a weak     inequality,  e.g. $3x_1 +   x_4 \leq 10$.  
  A     Horn arithmetic clause is a disjunction of any   number  of     disequalities
   and  at most one weak inequality. 
An example of a Horn arithmetic   clause is given below. Satisfiability of  Horn   arithmetic formulas  is polynomial \cite{KoubarakisS00}. 
Future work section considers the extension
of this class to NC form.

\vspace{-.3cm}
 $$(3x_1 + x_5 -4x_3 \leq 7) \vee (2x_1 + 3x_2 - 4x_3 \neq 4) \vee (x_2 + x_3 + x_5 \neq 7)$$
 
 \vspace{-.15cm}

\vspace{.15cm}  
\noindent $ \bullet $ {\em $\mathsf{Horn \ Constraint \ Systems.}$} 
Different classes  of numerical Horn  formulas are studied 
  in \cite{SubramaniW15}. The authors consider the
  next setting: if ${\bf  x}$  and ${\bf a}$ are vectors and  
  ${\bf A}$ is a matrix, a system of constraints ${\bf A \cdot x \geq a}$ is called 
  {\em Horn  constraint system} if: (a) the entries in {\bf A} are in $\{0, 1, -1\}$;
(b) each row of {\bf A} contains at most one positive entry; (c) {\bf x} is a real vector;
and (d) {\bf a} is a integral vector.
  A  comprehensive account of a variety of theses Horn constraint systems 
and their corresponding complexity is given in \cite{Wojciechowski19}. 
Whether or not these classes can be extended
to NC form is an open issue for us.

\vspace{.2cm}
\noindent $ \bullet $ {\em $\mathsf{\L ukasiewicz \ Logic.}$}
%
%
%
 The authors   in \cite{BorgwardtCP14}    defined Horn formulas in the 
infinite-valued \L ukasiewicz logic 
 and  proved that
their  satisfiability problem   is {\bf NP}-complete.  
 However,    it
was proven to be  polynomial   for  the    3-valued case \cite{BofillManyaVidalVillaret15b}.
 Further  tractable and intractable subclasses are given in \cite{BofillManyaVidalVillaret19}. Future work section
 regards this class to be extended to NC.

\subsubsection{Automated Reasoning} \label{subsec:Horn-reason}

  The models of  Horn formulas 
   form a lattice\footnote{A fact observed first  
   by McKinsey \cite{McKinsey43} (see also \cite{Horn51})}  and this 
   mathematical tool  is also the underlying  structure within a large number of domains: 
    lattice theory, hypergraph theory, 
     data bases, concept analysis,
   artificial intelligence, knowledge spaces,   Semantic Web, etc. This fact   entails that   
   many concepts and  algorithms  have been developed 
   for Horn formulas in different realms and
     there exist   bijective links between the  elements 
   belonging to different fields \cite{BertetM10}. 
   See \cite{AdarichevaIBT14,Wild17,BertetDVG18}    for
      comprehensive surveys.
  %
  Thus,  we give a short overview of some selected  reasoning problems 
  which    are
tighter connected with  $\mathbb{H_{NC}}$.       

\vspace{.15cm} 
 \noindent   {\bf Horn Functions.} 
Noting $\mathcal{B}=\{0, 1\}$, a Boolean function $f_\mathcal{B}$   is  a mapping 
  $\mathcal{B}^n \rightarrow \mathcal{B}$. 
$f_\mathcal{B}$ is modeled by a formula $\varphi$ {\em iff}  $f_\mathcal{B}$  
assumes  1 exactly on the 0-1 vectors corresponding to  the models of $\varphi$
and   0 elsewhere. 
The  next claims are well-known 
 \cite{HammerK92,HammerK93,daglib/0028067}.

\vspace{.1cm} 
$-  $ $f_\mathcal{B}$ is representable by numerous formulas.

\vspace{.cm}
$-  $ $f_\mathcal{B}$ is representable by a clausal formula. 

\vspace{.cm}
$-  $ $f_\mathcal{B}$ represents a unique Boolean function.

\vspace{.1cm} 
\noindent If $f_\mathcal{B}$ is a Horn function $f_\mathcal{H}$ 
and is   defined on $ (x_1, x_2, \ldots , x_n) $, then 
$f_\mathcal{H}$ fulfills  
  \cite{daglib/0028067}:

\vspace{.1cm} 
 $-  $ The set of models of $f_\mathcal{H}$ is closed under componentwise conjunction.

\vspace{.cm} 
$-  $  $f_\mathcal{H}$ is representable by a Horn formula.

\vspace{.cm}  
$-  $ The true points  of $f_\mathcal{H}$ form a closure system  on $\{x_1, x_2, \ldots , x_n \}$.

 \vspace{.2cm} 
\noindent Given a $f_\mathcal{H}$,   obtaining  its Horn formula  is done in \cite{Ben-KhalifaM09}.
    Horn functions are also representable by non-Horn formulas and 
     to decide whether  an arbitrary clausal formula represents a Horn function is co-NP-complete  \cite{daglib/0028067}.
A prime implicant,   fundamental   in  Boolean functions,  
     of $f_\mathcal{B}$ is a  clause $C$:    (i) whose  models 
    include those of  $f_\mathcal{B}$  
   and  (ii) no sub-clause of $C$ satisfies  (i). 
  Any   $f_\mathcal{B}$  is representable by the conjunction of its prime implicants.
   Searching all prime implicants of any  $f_\mathcal{B}$ is NP-complete.
   The    prime implicates 
   of any $f_\mathcal{H}$ are Horn clauses. 
  A main application of  Horn functions  is  finding the 
    Horn representation of any $f_\mathcal{H}$
   with   minimal size, which
      has several applications, e.g.
     compacting  knowledge bases  \cite{HammerK96}. 
 The steps to find the required prime implicants of any 
  $f_\mathcal{B}$  are: (1)
obtaining   the prime implicants of  $f_\mathcal{B}$  and (2)
     finding  which prime implicants can be discarded.   The minimization problem is   NP-hard, 
   for general and also for Horn funtions  \cite{HammerK93}. 
  Horn functions are a sub-class of other  more general 
   Boolean functions such as the renamable-Horn
 and the Q-Horn functions  \cite{daglib/0028067}
 representable by
 Horn-renamable  and Q-Horn formulas, respectively (see  Subsection \ref{subsec:Horn-Super-classes}).

\vspace{.0cm}
Given that  Horn functions \cite{daglib/0028067}  are representable by a Horn formula and Corollary 
 \ref{cor:firstcorprimer}  states that
 both  $\mathbb{H_{NC}}$ and the Horn  class are semantically
  equivalent, $\mathbb{H_{NC}}$ can serve   for the analysis of  Horn functions. In fact,  $\mathbb{H_{NC}}$
  turns out to be an attractive option to the Horn class taking into account that,
 as   proved in Section \ref{sec:semanticalsyntactical},   $\mathbb{H_{NC}}$ is  linearly recognizable  
and its HNC formulas   can  be even exponentially smaller 
 than their   clausal Horn representations. 

     \vspace{.15cm}  
 \noindent {\bf Implicational (Closure) Systems.}  They are used to 
express strong implications between attributes
  in databases (functional depdendencies) 
and relational data \cite{DechterPearl92}, but are relevant in many
 domains. We refer to \cite{AdarichevaIBT14,Wild17,BertetDVG18} 
 for   recent overviews.
{\em A formal context} $ \mathbb{K} $ is  
 a 3-tuple $(G,M,I)$, where $ G $ are objects, $ M $ are  attributes and $ I $ is a binary table
indicating which objects have  which of the attributes \cite{GanterWille99}.  
%
{\em A formal concept} is a 2-tuple $ (O_i, A_i) $   where $ O_i $ is a set of objects and $ A_i $ is a set of attributes and means that
every object in  $ O_i $ has  every attribute in $ A_i $. If an object  is not in
$ O_i $ then there is an attribute in $ A_i $ which that object does not have; and for every
attribute  that is not in  $ A_i $ there is an object in $ O_i $ that does not have that attribute.
An implication  $ A \rightarrow B $ means each object having all 
attributes in $ A $ also have all attributes in $ B $.  
{\em An implication system}  $ \Sigma $ defined on a set $ S $
is a binary relation on $ \mathcal{P} (S)$ and consists of a set of 
implication rules $ A \rightarrow B $, where $ A , B \in \mathcal{P} (S)$.
The family $ \mathcal{F} $ that satisfies all implication rules in $ \Sigma $ is:

\vspace{-.3cm}
$$ \mathcal{F}_\Sigma=\{X \subseteq S : A \subseteq X \  \mbox{imply} \ B \subseteq X 
\ \mbox{for all} \ A \rightarrow B \in \Sigma  \}$$

\vspace{-.15cm}
\noindent   $ (\mathcal{F}_\Sigma, \subseteq) $ is a closed set lattice 
where a closure of   $ X  \subseteq S $ is obtained via the operator $ \varphi_\Sigma $:

\vspace{.2cm}
$\bullet \  \varphi_\Sigma(X)=\pi(X) \cup \pi^2(X) \cup \pi^3(X) \cup \ldots$ (up to saturation) with

\vspace{.2cm}
$\bullet \  \pi(X)=X \cup \bigcup \,\{B : A \subseteq X \ \mbox{and} \ A \rightarrow B \in \Sigma \}$

\vspace{.1cm}
 
\noindent An algorithm   to obtain   $\varphi_\Sigma(X) $ is  in \cite{daglib/0028067}.
There are tight connections between   Horn functions and implication  systems
\cite{BertetM10}.
%
%
%
Implication systems can be viewed from
a logic-based perspective as follows. $ A \rightarrow B $ holds in $ \mathbb{K} $ if each object that belongs to attributes in $ A $
also belongs to atributes in $ B $. 
Any  $ \Sigma $ is complete for $ \mathbb{K} $ if, 
for all   $ A \rightarrow B $: \
  $ \mathbb{K} \models A \rightarrow B \ \ \mbox{iff} \ \   \Sigma \vdash A \rightarrow B$. 
%
 $ \Sigma $ and $ \Sigma' $ are equivalent if $ \varphi_\Sigma(X)=\varphi_{\Sigma'}(X)$.
  $ A \rightarrow B $ follows from $ \Sigma $, noted  $ \Sigma \models  A \rightarrow B$, 
if  $ \Sigma \cup  A \rightarrow B$ is equivalent to $ \Sigma $. Then,
we have   \cite{Wild17}: \ 
$ \Sigma \models A \rightarrow B \quad \mbox{iff} \quad B \subseteq  \varphi_\Sigma(A)$.
 The possibility  of managing implications by inference
systems is one of its outstanding features. 
Equivalent systems are obtained with  inference rules
whose aim  is to turn the   original system  into an equivalent system 
  fulfilling
some desired properties. 
Thus, many methods to compact
 implication system have been proposed  and  the first     was formed by the Armstrong  rules \cite{Armstrong74}.
 
 \vspace{.0cm}   
  Implication
systems   with the HNF syntax   can  be contemplated.  
Classical implications  can be extended  with HNF implications 
 as those of 
Definition \ref{def:NCrules} and Example \ref{ex:NCrule}. This would allow
for  compacter representations of   implication systems, which is one of their desired properties, viz.
optimality. However,     compacter syntactical forms  have  more complex structures and so 
defining inferences   with  HNF implications makes difficult to ensure 
  their completeness.  Another interesting   issue   is that of directness.
Directness with HNF implications amounts 
to, given any arbitrary  
set $ X $,  find  $\varphi_\Sigma(X)  $    with just one traversal of the HNF system. 
 A polynomial algorithm based on UR$_{\mbox{nc}}$  could be devised
 for getting $\varphi_\Sigma(X)  $  in an HNF direct implication-system,
 and likely even with linear complexity.

%

 \vspace{.2cm}
\noindent {\bf Directed Hypergraphs.} A directed hypergraph is an  pair $ (E,\mathcal{H}) $, $ E $ being 
 a vertex set  and $ \mathcal{H}  $ a   hyper-edge  set. 
Hypergraphs   have been applied in many domains including   implication systems.
   Horn formulas can be mapped into a
directed hypergraph e.g., \cite{AusielloL17} and
  Horn-SAT can be proved to be {\em P-complete       via 
P-completeness of the reachability 
problem in directed hypergraphs.}
 Assuming that a Horn formula is satisfiable and a new clause is inserted,
 the satisfiability question is answered in $O(1)$ time and inserting a new clause of 
 length $q$  in $O(q)$ amortized time \cite{AusielloI91}. 
If $m$ clauses of a unsatisfiable Horn formula are deleted, the  time needed 
to maintain the satisfiability state and its minimum model is $O(n\times m)$, 
$ n $ being the number of variables \cite{AusielloFFG97}.
%
%
We think that a reachability notion   can  be roughly defined 
 on the  DAG of any $\varphi \in \mathbb{H_{NC}} $  
as follows: (i) initially only positive-literal nodes in unit clauses are reachable; 
(ii) a negative-literal node is reachable if its dual positive-literal node is reachable;
(iii) a negative $\vee$-node (resp. $\wedge$-node)  is reachable if all  (resp. one) of its 
negative disjunct nodes are reachable;  
and (iv) the non-negative disjunct node in a $\vee$-node is reachable if all its negative disjunct nodes are reachable; and (v) all conjunct  nodes of a reachable non-negative $\wedge$-node are reachable. 
Thus $  \varphi$ is unsatisfiable iff   the node {\bf F} is reachable on the DAG of $  \varphi$.
The satisfiability of $ \mathbb{H_{NC}} $ also is  P-complete because
both it is polynomial and contains  Horn-SAT. The mentioned problems studied for the clausal
scenario when clauses are inserted or deleted from a (un)satisfiable  formula,
such as maintaining  the minimum model of an HNC, 
 are worthwhile issues to be tackled.
 
 \vspace{.2cm}
\noindent {\bf Horn Approximations.}    An alternative    to   knowledge compilation   has been developed in 
\cite{SelmanK91, KautzS92,SelmanK96, MakinoO12} and
 consists in  compiling a formula $ \varphi $ into two Horn formulas 
$H_{lb}$  and  $H_{ub}$ so that many queries performed under $ \varphi $ can be performed under $H_{lb}$  and  $H_{ub}$  polynomially.
 Thus, the goal is finding   $H_{lb}$  and  $H_{ub}$ whose models
$\mathcal{M}(H_{lb})$ and $\mathcal{M}(H_{ub})$ verify 
$\mathcal{M}(H_{lb}) \subseteq \mathcal{M}(\varphi) \subseteq \mathcal{M}(H_{ub})$. 
Then one disposes of: (i) sound approximation:   $ \mathcal{M}(H_{ub})   \models \phi$
implies $ \varphi   \models \phi$, and (ii) complete approximation:
$ \mathcal{M}(H_{lb}) \not \models \phi$
implies $ \varphi \not \models \phi$. To maximize these concepts, 
 the tightest approximations must be sought,
   which are: (i) $H_{glb}$  s.t.  
 $\mathcal{M}(H_{lb}) \subseteq \mathcal{M}(H_{glb}) \subseteq \mathcal{M}(\varphi)$
for all $H_{lb}$; and (ii) $H_{lub}$ s.t. $ \mathcal{M}(\varphi) \subseteq \mathcal{M}(H_{lub}) \subseteq \mathcal{M}(H_{ub})$
for all $H_{ub}$. Finding Horn approximations is mildly harder than solving the original clause entailment,
i.e.  
it is $\mathrm{ P^{NP[O(log n)] }}$-hard \cite{CadoliS00}. 
Determining  HNFs $ \varphi_1 $ and $ \varphi_2 $ verifying the requirements for    $H_{glb}$ and  $H_{lub}$
is clearly harder than determining the proper  $H_{glb}$ and  $H_{lub}$ as   clausal  formulas 
have simpler syntactical structure than    NC formulas.  However,  
some simple HNFs can be considered 
to start with, e.g. a  conjunction  of disjunctions formed by  negative literals 
and one  conjunction of positive literals,
which amounts to compact  Horn clauses sharing the same negative literals.
Besides, as  HNFs are much smaller than their Horn clausal counterparts and 
also clause entailment (unsatisfiability)
with HNFs is  polynomial, HNFs
will allow accelerating on-line answering. 

\vspace{.15cm}
\noindent {\bf Abduction.} Abduction is the problem of, given a Horn function $ f_H $
and a assumption set $ \mathcal{S} $ of propositions,
 generating an explanation for a literal $ \ell $,  which is a subset $ \mathcal{E} \subseteq \mathcal{S} $
verifying: (i)  $   f_H  \wedge \mathcal{E} $ is satisfiable, and (ii)  $   f_H \wedge \mathcal{E} \models \ell$.
The complexity of abduction depends on the representation of 
$ f_H $: if $ f_H $ is represented by its characteristic models, then abduction is polynomial \cite{KautzKS93},
wheras  it is NP-complete if $ f_H $  is represented by OBDDs \cite{HoriyamaI04}.
The complexity of abduction when Boolean functions are presented by HNC formulas
 should be studied, but  it is likely polynomial because the above 
 conditions (i) and (ii) are checked polynomially,  taking into account that $ \mathbb{H_{NC}} $  is polynomial 
 for satisfiability testing.
 
 \vspace{.15cm}
\noindent {\bf Petri Nets.} There are tight relations  between rule-based systems 
and Petri Nets   \cite{YangTC03,ChavarriaBaezXiaoouLi06,LiuLMZ13}.  
Petri Nets are used to verify properties of rule-based systems such as 
inconsistency, redundancy and circularity rules. A transition $ T $
having several input places $ P_1, P_2, \ldots, P_{j-1} $ and several output places 
$ P_j, P_{j+1}, \ldots, P_k $ is modeled by  rule (1) below:

\vspace*{.2cm}
(1) $ P_1 \wedge  \ldots \wedge P_{j-1} \longrightarrow  P_j   \wedge  \ldots \wedge  P_k$  \qquad
(2) $T_1 \vee  \ldots \vee T_{j-1} \longrightarrow T_j \wedge  \ldots \wedge T_k$

\vspace*{.2cm}
\noindent A place $ P $ preceded by several transitions  $ T_1, T_2, \ldots, T_{j-1}$ 
 and followed by several  transitions $ T_j, T_{j+1}, \ldots, T_k $ is modeled
 by rule (2) above.
 As  rules (1)-(2)  are not  Horn,
a normalization process  \cite{YangTC03}  must be applied, which of course significantly  increases the size of the obtained knowledge base.
However, both rules  are indeed  HNF  according to Definition  \ref{def:NCrules}, and so, 
they can be polynomially tackled  with   UR$_{\mbox{nc}} $-based inferences.

 \section{Future Work} \label{sec:futurework}
 
Our presented approach can be extended smoothly to 
heterogeneous logics,     accommodated to 
different reasoning settings, 
and used as a base towards   developing  the NC paradigm.   
Future research,   that is  likely to receive
  our attention, is divided into two axes.

\subsection{$\mathbb{H_{NC}}$ 
and        UR$_{\mbox{nc}}$ for Several Logics}  \label{subsect:HornAlg}
\vspace{.1cm}

We will define   $\mathbb{H_{NC}}$ 
and      UR$_{\mbox{nc}}$ for several logics
and validate each UR$_{\mbox{nc}}$   by proving its completeness 
for its corresponding $\mathbb{H_{NC}}$. 
Then, based on UR$_{\mbox{nc}}$, we will  design   
 Horn-NC-SAT algorithms, which are useful e.g. in DPLL reasoners.
%
%

\vspace{.15cm}
\noindent  {\bf Propositional logic.}    Since  $\mathbb{H_{NC}}$
should  play in NC   a   r\^ole  similar to the r\^ole played by   Horn    in   clausal efficiency, 
  worthy research efforts  remain  to  devise a highly-efficient Horn-NC-SAT algorithm. 
As   Horn-SAT  is linear, 
 proving   that  Horn-NC-SAT  is also linear   would suggest 
 that  reaching clausal efficiency
  in some NC    settings   is conceivable. 
We have  shown that  Horn-NC-SAT is polynomial and we  believe
that  its linearity can be proved.

\vspace{.15cm}
\noindent  {\bf  Logic Programming.}   This topic has been  
discussed in Subsection \ref{sub:applic-log-prorm}.

\vspace{.15cm}
\noindent    {\bf Quantified Boolean Formulas (QBFs).}   Building  
  solvers for QBFs in NC is a  contemporary research concern,    e.g.
    \cite{LonsingB08, EglySW09,BalabanovJ12}.
We will extend the presented    $\mathbb{H_{NC}}$   
to QBFs and  define the
QBF-Horn-NC  class.  Then,  we  will determine UR$_{\mbox{nc}}$ for QBF-NCs, basic for  DPLL. The complexity of QBF-Horn-NC is an open
question  knowing
that  QBF-Horn is quadratic \cite{BuningKullmann09c}.

\vspace{.15cm}
\noindent   {\bf   Regular  Many-Valued Logic} and 
 {\bf     Possibilistic  Logic.}  
   Our approach has been extended 
   to these logics in  \cite{Imaz21b,Imaz2021c}.

\vspace{.15cm}
\noindent  {\bf Linear-Arithmetic Logic.} In Related Work, Arithmetic-Horn formulas
have been presented and mentioned that 
its associated SAT problem  is polynomial \cite{KoubarakisS00}.  
 Thus, we will  define the Arithmetic-Horn-NC class.
 Then,    we will try to prove that Arithmetic-Horn-NC-SAT  is polynomial 
  as its clausal counterpart   is \cite{KoubarakisS00}. For that purpose, we will require first
  to stipulate UR$_{\mbox{nc}}$  for linear-arithmetic logic.

\vspace{.15cm}
\noindent  {\bf \L ukasiewicz Logic.}  
As evoked in Related Work, the satisfiability problem of the class $\mathcal{L}_{\infty}$-Horn is {\bf NP}-complete \cite{BorgwardtCP14} and
   polynomial \cite{BofillManyaVidalVillaret15b}  for  the    3-valued case, $\mathcal{L}_3$-Horn.
 Our  goal will be  to lift 
  $\mathcal{L}_3$-Horn 
   to NC,   
%
   determining thus, the $\mathcal{L}_3$-Horn-NC class,  $\mathcal{L}_3$-$\mathbb{H_{NC}}$.
 Then we will analyze whether tractability is preserved in  NC, that is,
 whether  $\mathcal{L}_3$-$\mathbb{H_{NC}}$ is polynomial. For that, a former step 
  is to define 
  UR$_{\mbox{nc}}$ for $\mathcal{L}_3$. Posteriorly, we will deal with the infinite-valued case.


\subsection{Developing the NC Paradigm} \label{subsect:DPLL solvers}

\noindent{\bf  Tractability.}
This research line   will  be  resumed  
 by   uplifting    some  of the polynomial Horn-clausal extensions    reviewed 
 in Related Work. Concretely,  we will first attempt to determine the 
renamable-Horn-NC and  SLUR-NC classes.
 Indeed, uplifting polynomial clausal classes  may be an open research direction
    towards amplifying NC tractability.

\vspace{.2cm}
\noindent   {\bf Satisfiability.}   It was empirically verified (Related Work) 
    that  DPLL outperforms the other  published methods. $\mathbb{H_{NC}}$ and  UR$_{\mbox{nc}}$   are the base
of  NC DPLL solvers.
Hence,  
 we plan: (1)
    devising  a fast  Unit-Propagation 
based on UR$_{\mbox{nc}}$
  and easing its suitable integration       
  into  the NC-DPLL structure; and (2)   lifting     polynomial  classes  surveyed in  Related Work to NC form.


\vspace*{.1cm}  
\noindent    {\bf   MaxSAT.}
As evoked  in Related Work, NC MaxSAT has not been properly defined. 
At least  the next three optimization factors  can be considered,   $\varphi$ being the input:
(i) the number   of    satisfiable clauses in  
   $cl(\varphi)$; (ii) the number   of    satisfiable sub-formulas in  
   $\varphi$; and (iii) a  weighted  factor  
   resulting from combining both  previous factors.
We will analyze such variants, compare them and solve  NC-MaxSAT.

\vspace{.2cm}
\noindent  {\bf  Signed Many-Valued Logic.} 
  Polarity  of literals 
 is the underlying 
  characterization  of  the   Horn   pattern;  
   yet, it does not apply in signed many-valued logic \cite{BeckertHahnleManya00,Hahnle01}.
However,   a realistic possibility to make tractability appear  in signed logic 
  consists in extending     polynomial
Horn  super-classes   
 to signed  logic. We think that  
the simplicity, wide range and relevance of the {\em SLUR formulas} (Related Work)  make them     good candidates to begin with. 
%
Then, we will attempt to lift 
the   found   clausal  subclasses to NC.

\vspace{.2cm}  
\noindent    {\bf  Resolution.}  N. Murray in the 1980s \cite{Murray82}  
proposed NC Resolution   
    in a combined functional and classical manner.
  Thus,   we plan to make the next contributions: 
  (1)  determining   a new definition 
 of  NC Resolution,    
 making it  as  deterministic as  clausal Resolution is,
 and proving also its   refutation completeness. 
 We will pursue this objective
 following the  principle employed  in this article which led us to define UR$_{\mbox{nc}}$; and 
  (2)     NC Hyper-Resolution 
has  not been defined yet, and thus we will attempt to bridge such gap.

\vspace{.2cm}  
\noindent {\bf  Q-Resolution.}  Resolution   QBFs  is called Q-Resolution \cite{DBLP:journals/iandc/BuningKF95}.
Thus, we plan   lifting  
 Q-Resolution to  NC and so defining NC-Q-Resolution.
 Yet, in view of the  aforementioned  difficulties    to prove 
 the completeness  of (non-quantified)  NC-Resolution,    
  it will be mandatory to start with 
 simple  conjunctions of hybrid CNF  and DNF instances
   of QBF-NCs.

\vspace*{.2cm}  
\noindent   {\bf   Theorem Proving.}
  Future research pointed out so far,  deals  with propositional  
  logic and with some of its extensions.  As  long-term research project, 
  we plan to enlarge   the scope of our approach to 
   first-order logic (FOL).
%
%
   Nonetheless,   attending to the complexity
  of  reasoning  with NC and FOL,    we will begin 
   with simple fragments   such as those limited to  monadic and binary predicates 
   and  having  at most one  function symbol.
   %


\section{Conclusion} \label{sec:futureWo}

Our first main contribution has been  the  
   definition of the 
hybrid   Horn-NC class,  $\mathbb{H_{NC}}$,
  obtained by adequately amalgamating both
       Horn  and   NC classes, or equivalently, by lifting the  Horn  pattern to NC.  
           We  first  defined 
   $\mathbb{H_{NC}}$ as the class of NCs whose disjunctions have at most
 one non-negative disjunct, and then, provided a syntactically  detailed, compact,
 inductive definition of $\mathbb{H_{NC}}$. Furthermore, we proved its next properties:
\begin{enumerate}

 \item $\mathbb{H_{NC}}$ is polynomial for satisfiability testing.
 
 \vspace{-.25cm}
 \item  $\mathbb{H_{NC}}$  syntactically subsumes $\mathbb{H}$.
  \vspace{-.25cm}
 \item  $\mathbb{H_{NC}}$ and $\mathbb{H}$ are   semantically equivalent.
  \vspace{-.25cm}
 \item $\mathbb{H_{NC}}$ contains all  NCs  whose clausal form  is Horn.

 \vspace{-.25cm}
 \item $\mathbb{H_{NC}}$  is linearly recognizable.
  \vspace{-.25cm}
 \item $\mathbb{H_{NC}}$ is   strictly more succinct   than $\mathbb{H}$.
  \vspace{-.25cm} 
\item $\mathbb{H_{NC}}$ is    exponentially richer  than $\mathbb{H}$.
\end{enumerate}

Our second main contribution has been  establishing the calculus Non-Clausal Unit-Resolution, UR$_{\mbox{nc}}$,  and proving that 
it   checks the satisfiability of  $\mathbb{H_{NC}}$ in polynomial time, which   makes $\mathbb{H_{NC}}$
the first  determined tractable class in NC reasoning. 

\vspace{.05cm}
The advantages of $\mathbb{H_{NC}}$ and UR$_{\mbox{nc}}$
include its clear potential to leverage  NC reasoning: (i) $\mathbb{H_{NC}}$ and UR$_{\mbox{nc}}$
are crucial to build   NC DPLL satisfiability solvers and 
NC DPLL theorem provers capable of emulating the efficiency of their clausal counterparts; and
(ii) a general NC reasoner supplied with UR$_{\mbox{nc}}$ can 
polynomially decide  the HNC fragment.

\vspace{.05cm}
Also, we have highlighted that declarative languages and in general rule-based systems
may enrich their languages considering NC formulas in their antecedents and consequents
and which can be done with a query-answering efficiency  comparable to
 clausal efficiency.
As a by-product of the  properties of $\mathbb{H_{NC}}$,  it could draw interest from other fields 
such as    knowledge compilation,     Horn functions analysis or   implication systems.

\vspace{.05cm}
We plan  to follow several future research directions: 
 proving that   satisfiability checking $\mathbb{H_{NC}}$
is  linear; 
   exploring the  scope of $\mathbb{H_{NC}}$ and UR$_{\mbox{nc}}$
  in  NC logic programming;
  defining  $\mathbb{H_{NC}}$ and  UR$_{\mbox{nc}}$ beyond propositional logic;
   founding new polynomial NC classes ;
 obtaining a clausal-like definition of Non-Clausal Resolution; 
and   developing the NC paradigm through a number of outlined ways.



 \vspace{.2cm}
\noindent {\bf Acknowledgments} J. Buenabad-Ch\'avez, Llu\'is Godo.

\end{document}